\documentclass{article}
\usepackage{amsthm}

\newif\ifanon
\anonfalse

\ifanon\else\PassOptionsToPackage{super,comma,sort&compress}{natbib}\fi
\usepackage{iclr2027_conference,times}

\ifanon\else\iclrfinalcopy\fi

\usepackage[utf8]{inputenc} %
\usepackage[T1]{fontenc}    %
\usepackage[colorlinks,allcolors=black]{hyperref}       %
\usepackage{url}            %
\usepackage{booktabs}       %
\usepackage{amsfonts}       %
\usepackage{amssymb}        %
\usepackage{microtype}      %
\usepackage[table]{xcolor}         %

\usepackage{wrapfig}

\usepackage{graphicx}
\usepackage{amsmath} 
\usepackage{mathtools}
\usepackage{bm}

\usepackage{amsmath,amsfonts,bm}

\def\eqref#1{equation~\ref{#1}}

\def\1{\bm{1}}

\def\rvf{{\mathbf{f}}}
\def\rvg{{\mathbf{g}}}
\def\rvh{{\mathbf{h}}}

\def\rvp{{\mathbf{p}}}

\def\rvr{{\mathbf{r}}}

\def\va{{\bm{a}}}

\def\vs{{\bm{s}}}

\def\vx{{\bm{x}}}
\def\vy{{\bm{y}}}
\def\vz{{\bm{z}}}

\def\mS{{\bm{S}}}

\def\mZ{{\bm{Z}}}

\DeclareMathAlphabet{\mathsfit}{\encodingdefault}{\sfdefault}{m}{sl}
\SetMathAlphabet{\mathsfit}{bold}{\encodingdefault}{\sfdefault}{bx}{n}

\def\gA{{\mathcal{A}}}

\def\gS{{\mathcal{S}}}

\def\gZ{{\mathcal{Z}}}

\newcommand{\E}{\mathbb{E}}

\usepackage{float}
\usepackage{multirow}
\usepackage{tabularx}
\usepackage{etoc}
\usepackage{tikz}
\usetikzlibrary{arrows.meta, positioning}
 
\usepackage{booktabs}
\usepackage{tabularx}
\usepackage{subcaption}

\usepackage{pifont}

\usepackage{fontawesome5}
\definecolor{badgeink}{HTML}{1A1A1A}
\definecolor{badgeborder}{HTML}{D0D0D0}
\definecolor{badgefll}{HTML}{F4F4F4}
\newcommand{\paperbadge}[1]{%
  \begin{tikzpicture}[baseline=(L.base)]
    \node[inner xsep=8.2pt,inner ysep=4.8pt,rounded corners=4pt,
          draw=badgeborder,fill=badgefll,line width=0.45pt,
          font=\normalsize\sffamily,text=badgeink] (L) {#1};
  \end{tikzpicture}%
}

\usepackage[most]{tcolorbox}
\newtcbtheorem{definition}{Definition}{%
  enhanced,
  colback=gray!4,
  colframe=gray!40,
  fonttitle=\bfseries,
  coltitle=black,
  colbacktitle=gray!35,
  attach boxed title to top*=,
  boxed title style={colframe=gray!40, colback=gray!35, boxrule=0pt},
  separator sign={:~},
  left=5pt, right=5pt, top=3pt, bottom=3pt,
  before skip=8pt, after skip=8pt,
}{dfn}

\usepackage{caption}
\usepackage{enumitem}
\setlist{itemsep=1pt, topsep=2pt}

\usepackage[section]{placeins}

\title{Understanding the Limits of Latent World Models}
\author{%
  Rodrigo Gonzalez Laiz \& Steffen Schneider\\[.5em]
  Helmholtz Munich \\
  Munich Center for Machine Learning (MCML)
}

\begin{document}

\maketitle
\ifanon\else\lhead{}\fi

\begin{abstract}
World models are widely proposed as core components for enabling agents to reason over long temporal and spatial horizons. Yet, current evaluations often emphasize task performance while providing limited insight into whether learned world models acquire predictive latent dynamics that reflect the underlying structure of the environment. Here we introduce the concept of faithful latent world models as an equivariance condition between the environment's transition map and the learned latent dynamics. We propose latent rollout consistency (LRC) as a metric to evaluate action-conditioned latent prediction. Strikingly, we find that across the studied state of the art latent world models, none solves the task by actually faithfully modeling the dynamics of the environment.
\ifanon\else
\begin{center}
\vspace{0.15em}
\href{https://github.com/dynamical-inference/faithful-world-models}{\paperbadge{\raisebox{-0.12ex}{\faCode}\, Code}}%
\hspace{0.7em}%
\href{https://worldmodels.science/analysis}
{\paperbadge{\raisebox{-0.12ex}{\faGlobeAmericas}\, Website}}%
\end{center}
\fi
\end{abstract}

\ifanon
{\renewcommand{\thefootnote}{}\footnotetext{Code available at \url{https://anonymous.4open.science/r/world-models-37ED/}}}
\fi

\etocdepthtag.toc{main}
\section{Introduction}

Building agents that can reason over extended temporal and spatial scales is a major frontier in AI.
It has been argued that general-purpose agents need \emph{world models} if they are to accomplish complex, goal-directed tasks~\citep{ha2018worldmodels,lecun2022path,richens2025general}.
To support such behavior, world models must predict how the environment evolves under the agent's actions. However, latent world models are still evaluated primarily through task reward. Although reward measures whether the overall agent succeeds, it does not reveal whether the learned latent dynamics capture the underlying dynamics of the environment. 

This motivates the central question of this work: \emph{Do successful latent world models actually learn representations and dynamics that model their environment?} Recent work has identified failures of physical understanding in related settings. For example, \citet{joseph2026interpreting} study passive video encoders such as V-JEPA and VideoMAE through probing and representation geometry but do not evaluate future rollouts. \citet{motamed2026generative} evaluate long-horizon observational video prediction from visual context using metrics in pixel space. In contrast, we study \textit{action-conditioned dynamics} in agents that interact with and control an environment.
We evaluate DreamerV3, TD-MPC2, and LeWM~\citep{hafner2025mastering,hansentd,leworldmodel2026} on the Pendulum, CartPole, and Reacher environments in MuJoCo. These environments are \emph{deliberately simple} and serve as minimal test cases for world modeling. Their low-dimensional, well-defined dynamics allow us to isolate whether the learned world models accurately capture the underlying environment dynamics. Failures in these controlled settings therefore provide counterexamples and let us isolate areas of possible improvement to guide development of future algorithms.

For evaluation, we introduce the concept of \emph{faithful world modeling}. A world model is faithful if its representation preserves the environment state and its latent dynamics evolve consistently with the true dynamics under the same actions. We use latent rollout consistency (LRC) and post-hoc state decoding as metrics to measure this property.
Our results reveal two important limitations of current world models and the common evaluation protocols.
First, task reward alone does not reveal whether a world model learns accurate latent dynamics or preserves the underlying physical state. LRC and state decoding expose these failures directly. Second, these diagnostics reveal two recurring failure modes under pixel observations: inaccurate velocity estimation and difficulty learning dynamics through nonlinear observation mappings. Together, our findings suggest that improving both learned representations and latent dynamics is central to building more faithful world models.

\section{What does it take to model the world?} \label{sec:theory}

\begin{figure}[t]
    \centering
    \includegraphics[width=0.9\linewidth]{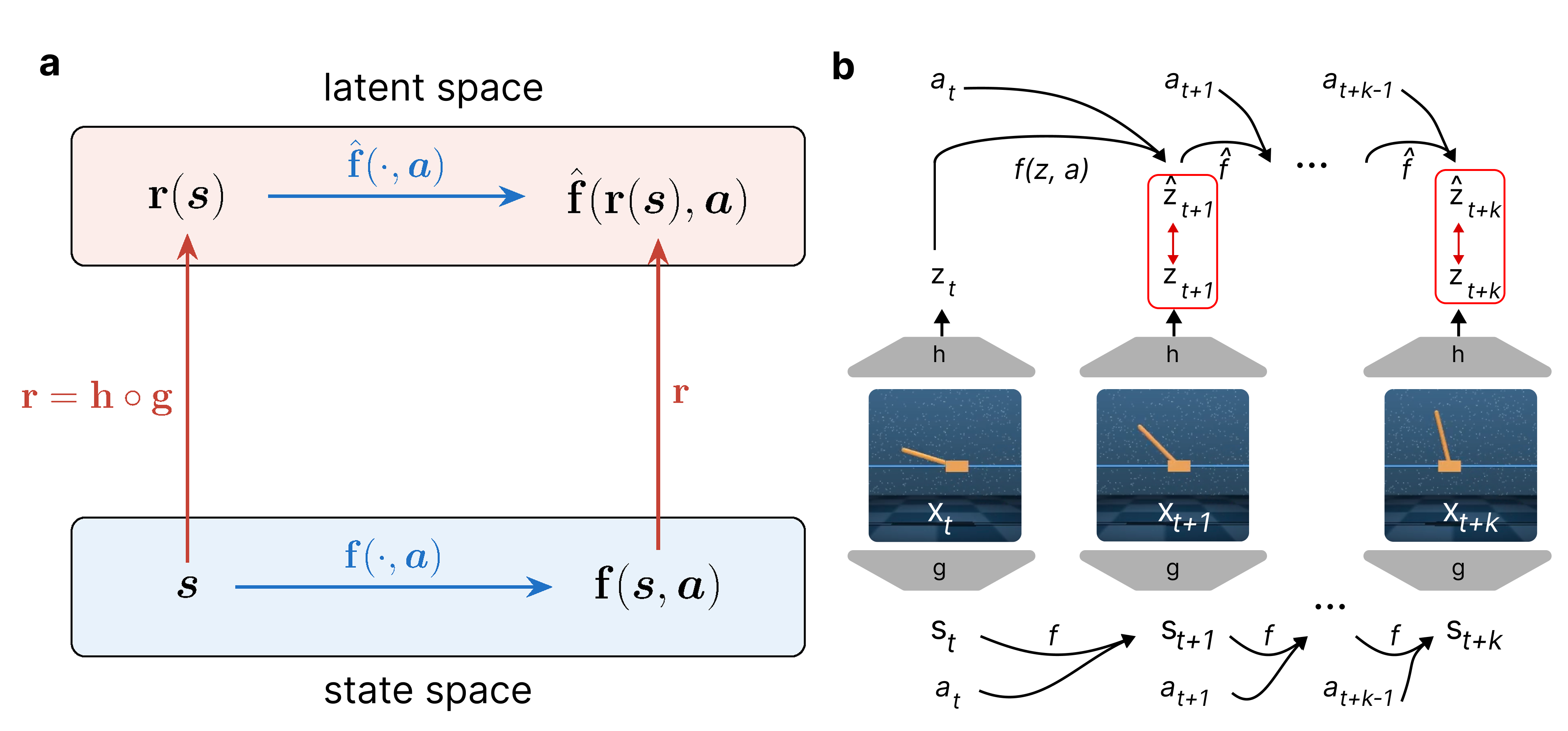}
    \caption{\textbf{Faithful latent dynamics and their evaluation.}
    (a) In a faithful latent world model, applying the true transition $\rvf$ and then encoding ($\rvr$) gives the same result as encoding first and applying the learned latent dynamics $\hat{\rvf}$ under the same action.
    (b) Latent rollout consistency (LRC) evaluates this by comparing autoregressive latent predictions $\hat{\vz}_{t+k}$ with representations $\vz_{t+k}$ inferred from the corresponding realized observations.}
    \label{fig:overview}
\end{figure}

We model the environment as a partially observed controlled dynamical system. The considered environments have deterministic dynamics and observations,  
\begin{equation}
\begin{aligned}
  \vs_{t+1} = \rvf(\vs_t,\va_t), \qquad
  \vx_t = \rvg(\vs_t),
\end{aligned}
\end{equation}
where $\vs_t$ is the unobserved environment state, $\va_t$ the action, and $\vx_t$ the observation. The observation may be a nonlinear view of the state. Following nonlinear independent component analysis, we call $\rvg$ the mixing function \citep{Hyvarinen2023-ua}.
A \emph{latent} world model has two coupled roles: (1) infer, or ``demix,'' a latent representation from observations, and (2) model how that representation evolves under actions. This gives the following definition of a latent world model:
\begin{align}
  \vz_t = \rvh(\vx_{\leq t},\va_{<t}) \qquad
  \hat{\vz}_{t+1} = \hat{\rvf}(\vz_t, \va_t),
\end{align}
where $\rvh$ is the learned encoder and $\hat{\rvf}$ is the learned dynamics model. We use the term specifically for models that infer the action-conditioned latent dynamics governing an environment. Generative video models \citep{ho2022video} and cognitive/spatial maps \citep{o1978hippocampus} are often forms of ``world models'' but are conceptually very different from the world models studied here, and hence, are out of scope for this analysis.

We are interested in the question: \textit{when is a world model $(\hat{\rvf},\rvh)$ a good model of the world $(\rvf,\rvg)$?} Two conditions follow naturally. First, the model should recover all information about the state from the observations. This condition is task-independent and allows the same world model to be used for different tasks, as long as those tasks are defined within the given state space. Second, predicting the evolution of an encoded state should agree with encoding the state after a true transition. This ensures that latent rollouts follow the dynamics of the environment. We define: %

\begin{definition}{Faithful latent world model}{faithfulwm-def}
Consider an environment with states $\vs \in \gS$, actions $\va \in \gA$,
transitions $\vs_{t+1} = \rvf(\vs_t, \va_t)$ and observations $\vx_t = \rvg(\vs_t)$
given by a mixing function $\rvg$.
A latent world model $(\rvh, \hat{\rvf})$ consists of an encoder $\rvh$ into a latent space $\gZ$ and a latent dynamics model
$\hat{\rvf} : \gZ \times \gA \to \gZ$.
An initial state $\vs_0 \in \gS$ and actions $\va_0, \va_1, \ldots \in \gA$
determine states $\vs_t$ and, by the mixing followed by the encoder, latents
\begin{equation}
  \vz_t = \rvh\big(\rvg(\vs_0), \ldots, \rvg(\vs_t), \va_{<t}\big) \in \gZ.
\end{equation}
The latent world model is \emph{faithful} if there is a context length $k_0 \in \mathbb{N}$ such that,
for any two such trajectories $(\vs_t, \vz_t)$ and $(\vs'_t, \vz'_t)$ and all $t, t' \ge k_0$:
\begin{enumerate}
  \item \label{eqn:faithfulwm}\emph{Dynamics equivariance:}\ \ $\hat{\rvf}(\vz_t, \va_t) = \vz_{t+1}$.
  \item \emph{State sufficiency:}\ \ $\vz_t = \vz'_{t'}$ implies $\vs_t = \vs'_{t'}$.
\end{enumerate}
\end{definition}

The first condition establishes an equivariance relation between the environment and latent dynamics: encoding after a true transition must agree with predicting in latent space. This is closely related to MDP homomorphisms \citep{ravindran2002model}, which preserve transition and reward structure, and to the consistency requirements of causal abstractions \citep{rubenstein2017causal,beckers2020approximate}. Unlike these frameworks, which allow information-losing abstractions, we focus on task-independent dynamics and additionally require injectivity to preserve the full environment state.

\paragraph{Learning the equivariance condition.}
The equivariance condition can, in principle, be optimized from observed transitions without access to the true state. Along any trajectory, the encoded next history $\vz_{t+1}=\rvh(\vx_{\leq t+1},\va_{<t+1})$ provides the target for $\hat{\rvf}(\vz_t,\va_t)$.
Many approaches, therefore, match the predicted latent state with the encoded next observation,
\begin{equation}
  \min_{\rvh,\hat{\rvf}}\;
  \E\!\left[
  \left\|\hat{\rvf}(\rvh(\vx_{\leq t},\va_{<t}),\va_t)
  -\rvh(\vx_{\leq t+1},\va_{<t+1})\right\|^2\right]
  +\lambda\Omega(\rvh,\hat{\rvf}).
\end{equation}

Here, $\Omega$ broadly denotes additional learning signals used to prevent collapse. These include reconstruction losses \citep{hafner2025mastering,hafner2018planet}, contrastive learning \citep{laiz_self-supervised_2024}, regularization \citep{balestriero2025lejepa}, and, in reward-based models, reward and value prediction \citep{hafner2025mastering,hansentd}. %

\paragraph{Metrics for faithful world modeling.}
Definition~\ref{dfn:faithfulwm-def} gives two conditions that we evaluate separately. We quantify empirical violations of dynamics equivariance using latent rollout consistency (LRC) and assess violations of injectivity using post-hoc state decoders.
At rollout horizon $H$, we compare the autoregressively predicted latent $\hat{\vz}_H$ with the latent inferred from the realized trajectory:
\begin{equation}\label{eq:lrc} 
  \mathrm{LRC}(H)
  =R^2\!\left(\rvh(\vx_{\leq H},\va_{<H}),\hat{\vz}_H\right).
\end{equation}
Appendix~\ref{app:lrc-validation} motivates the choice of $R^2$ and validates LRC against prediction accuracy in physical space.
A faithful model has $\mathrm{LRC}(H)=100\%$ at every horizon. High LRC, however, is not sufficient on its own: model collapse can still produce latents that are easy to predict. We therefore use post-hoc state decoding to test how much of the underlying state is found in the representation. Good decoding performance supports state sufficiency but does not guarantee injectivity.

\paragraph{Violations of the faithful condition.}

\begin{wrapfigure}[22]{r}{0.45\linewidth}
  \centering
  \vspace{-1.4em}
  \includegraphics[width=\linewidth]{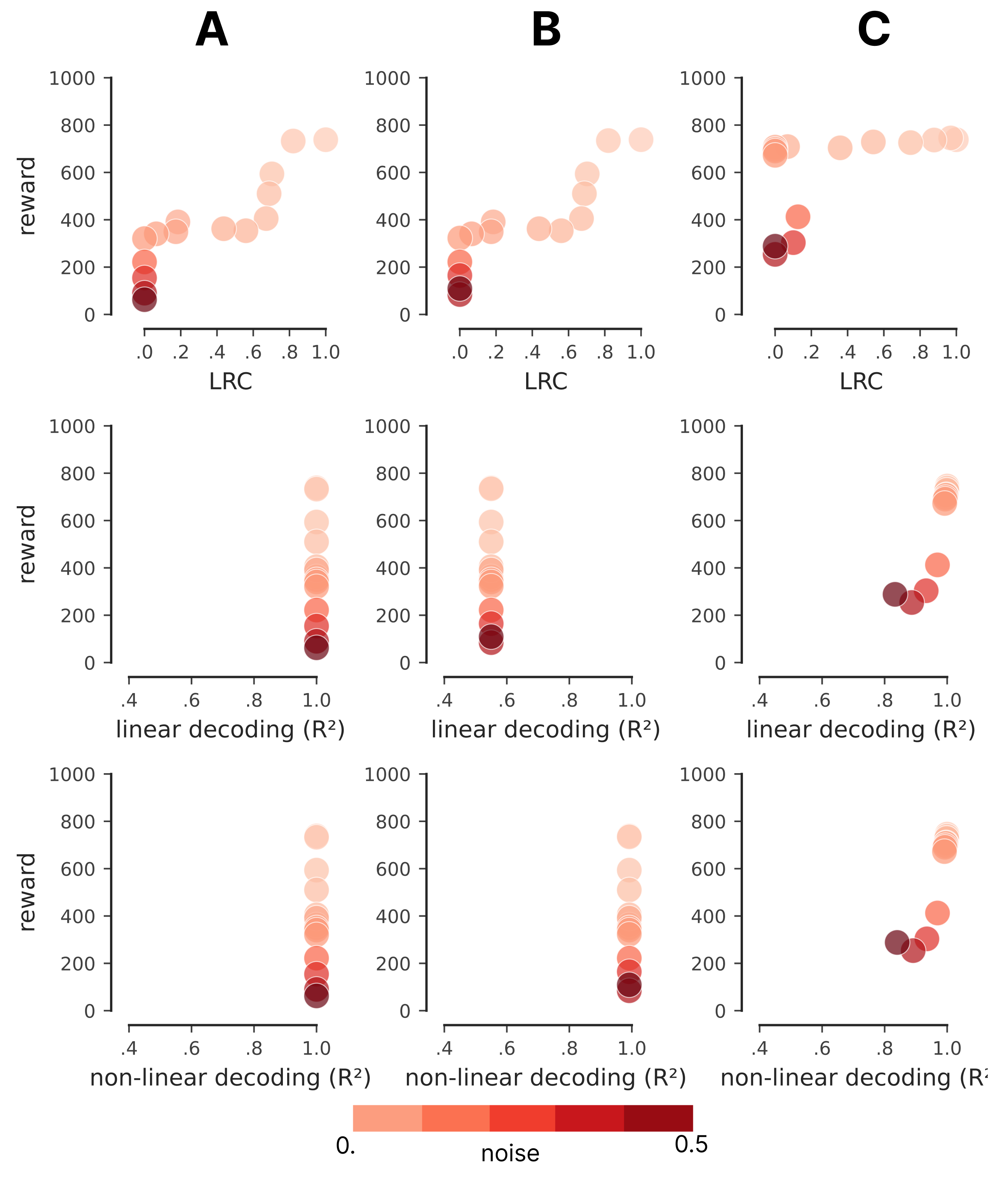}
  \caption{\textbf{A.}~identity encoding, noisy dynamics;
  \textbf{B.}~nonlinear encoding, noisy dynamics;
  \textbf{C.}~noisy encoder, exact dynamics.
  LRC at $H=30$, linear $R^2$, nonlinear $R^2$.
  }
  \label{fig:hypotheses}
\end{wrapfigure}

From left to right, Figure~\ref{fig:hypotheses} shows three ways a world model can fail to be faithful, using CartPole swingup with MPC and the true simulator as the planning model (Appendix~\ref{app:faithful-perturbations}).

Column A uses an identity encoding and adds noise to the oracle transitions. Linear and nonlinear decoding stay at $R^2=1$, so probes cannot see the failure, while reward falls with LRC.
Column B applies the same dynamics noise after an invertible nonlinear mix. Linear decoding is already poor at zero noise, because the coordinates are nonlinear, whereas nonlinear decoding stays at 1. Reward again tracks LRC rather than linear $R^2$.
Column C keeps the transitions exact and corrupts the inverse encoder. Decoding remains high and reward can stay high even as LRC collapses: encoder error is amplified over the rollout before return or probes break.

These settings show that both the representation and the dynamics model can limit task performance. They also motivate evaluating latent prediction, state decoding, and reward together.

\newpage
\section{Related Work}
\label{sec:related-work} 

\paragraph{World models and model-based RL.}
    The role of explicit internal models in intelligent behaviour has been debated since~\citet{brooks1991intelligence}, with \citet{richens2025general} recently formalizing the opposing view that any sufficiently general goal-directed agent must implicitly learn one. The modern formulation of \emph{world models} as compact latent dynamics models learned by an agent from its own experience was popularized by~\citet{ha2018worldmodels}. Contemporary world models span video generators (surveyed in~\citet{zhu2024sora}), JEPA-style latent predictors~\citep{agrawal2016learning,zhou2411dino,sobal2025learning}, including LeWM~\citep{leworldmodel2026}, and task-driven model-based RL agents---PlaNet~\citep{hafner2018planet},  DreamerV3~\citep{hafner2025mastering}, and TD-MPC2~\citep{hansentd}; diffusion-based variants such as DIAMOND~\citep{alonso2024diffusion} replace the latent dynamics with a generative model. Several recent works question whether such models actually capture environment physics: \citet{motamed2026generative} show that high-fidelity video models violate basic physical principles; \citet{garrido2025intuitive} evaluate V-JEPA with violation-of-expectation tests on IntPhys, GRASP, and InfLevel; and \citet{joseph2026interpreting} use layerwise probing, subspace analysis, and targeted ablations to study physical representations in V-JEPA~2 and VideoMAE-v2. Closest in spirit, \citet{ying2025assessing} call for metrics that quantify how well an agent has inferred the latent dynamics beyond task success.

\paragraph{Probing learned representations.}
    We build on the idea of probing classifiers~\citep{belinkov2022probing}. Probes have revealed emergent latent state in Othello-GPT~\citep{li2023emergent,nanda2023emergent} and have been used to reverse-engineer trained RL agents, from DQN~\citep{zahavy2016graying} and AlphaZero~\citep{mcgrath2022acquisition} to recent model-free agents and recurrent world models~\citep{bush2025interpreting,taufeeque2025path,simmons2025deep,sobotka2025reverse}. A complementary line probes the physical structure in foundation models~\citep{vafa2025has,liu2026kepler}. We apply this idea to state-of-the-art world models to ask whether reliable physical content is recoverable in the first place.

\paragraph{Identifiable representation learning.}
    Our analysis connects to identifiable representation learning, which studies when the underlying data-generating process can be provably recovered from observations~\citep{zimmermann2021cl,mikulasch2026understanding}, with foundations in nonlinear ICA~\citep{hyvarinen2019nonlinear,Hyvarinen2023-ua} and has recently been studied in the context of latent world models~\citep{laiz_self-supervised_2024}.
    Recently, \citet{reizingerposition} asked for more empirically grounded identifiability theory in self-supervised learning, a position our results reinforce: current world-model objectives fail to invert the data-generating process and learn faithful environment dynamics. This might be due mismatches in the assumptions of theoretically-motivated papers, or in properties of the numerical implementations that do not fully match the theory.

\paragraph{Time-series modeling and system identification.}
    Learning dynamics from sequential data has a long history in classical state-space models and filtering~\citep{kalman1960new,durbin2012time}, system identification~\citep{ljung1999system,vanoverschee1994n4sid}, and deep learning approaches to forecasting~\citep{rangapuram2018deep,salinas2020deepar,nie2023time}. Predictive state representations~\citep{littman2001predictive,boots2011closing} and Koopman-based methods~\citep{lusch2018deep} learn latent representations of dynamical systems in which the dynamics take a simpler form. These methods are typically evaluated by predictive accuracy in observation space, while identifiability guarantees from system identification rely on assumptions such as linear dynamics that do not carry over to learned latent world models. Our faithfulness condition instead evaluates whether the learned latent dynamics are consistent with the true environment transitions.

\section{Experimental setup}

\begin{figure}[t]
  \centering
  \includegraphics[width=\linewidth]{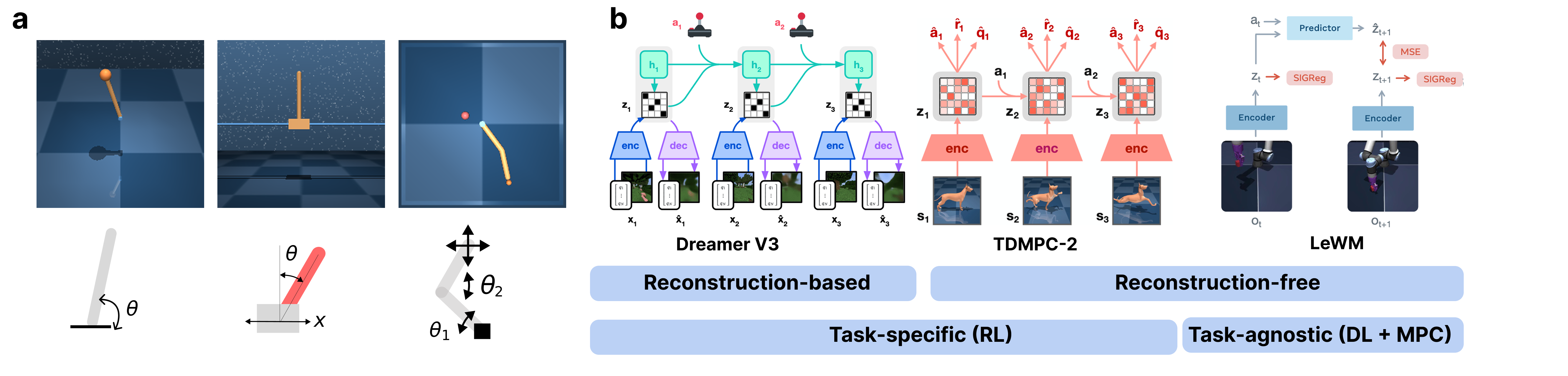}
  \caption{\textbf{Experimental setup.} (a)~Three world model families: TD-MPC2 and DreamerV3 (task-specific) and LeWM (dynamics learning + MPC), grouped by reconstruction and training objective. (b)~Three DeepMind Control Suite environments---Pendulum, CartPole, and Reacher---evaluated with state and pixel observations.}
  \label{fig:experimental-setup}
\end{figure}

\subsection{Benchmarks and training}
\label{sec:data}

We evaluate three complementary world-model families spanning two objective
axes (Figure~\ref{fig:experimental-setup}a) on three DeepMind Control Suite~\citep{tunyasuvunakool2020} environments (Figure~\ref{fig:experimental-setup}b). DreamerV3 is reconstruction-based
and reward-driven/task-specific, TD-MPC2 is reconstruction-free but reward-driven/task-specific, and LeWM is reconstruction-free and reward-free/task-agnostic, learning through a self-supervised latent-prediction objective and performing model-predictive control (MPC) as a post-hoc step.

We use three low-dimensional continuous-control environments: Pendulum
($d_s{=}3$), CartPole ($d_s{=}5$), and Reacher ($d_s{=}6$). Pendulum and
CartPole exhibit nontrivial unforced dynamics, whereas Reacher remains
stationary from rest without control; Appendix~\ref{app:envs_data} provides further details. In all three cases, we have access to
the ground-truth simulator state and dynamics, allowing us to evaluate what
the models learn rather than relying on task reward alone
\citep{posner2026observation}. These three environments are not meant to form a comprehensive benchmark. They are deliberately simple test cases. Doing well on them does not imply that a model will scale to harder settings, but if it cannot learn accurate dynamics here, it is difficult to expect it to do so in more complex systems.

We train each model family on both raw state vectors and rendered RGB images
($64{\times}64$~px for TD-MPC2 and DreamerV3; $224{\times}224$~px for LeWM).
Because the original LeWM implementation does not support state inputs, we
implement a state-based variant to enable a direct comparison with the other
models and with pixel-based LeWM. We train four seeds per model, modality, and
environment. Implementation and training details are provided in
Appendix~\ref{app:models}. 

\subsection{Evaluation protocol}
\label{sec:eval-protocol}

For each environment, we collect a separate evaluation set of 1,000 100-step episodes using a random Ornstein--Uhlenbeck (OU) policy designed to provide broad state-space coverage. Appendix~\ref{app:envs_data} describes the data-collection procedure and compares the state, action, and reward distributions of the collected datasets in Figures~\ref{fig:data-distribution-pendulum}--\ref{fig:data-distribution-reacher}.

\textbf{Latent prediction (LRC).}
\label{sec:eval-lrc}
We compute LRC (Eq. \ref{eq:lrc}) from one autoregressive latent rollout initialized at the beginning of each episode. Each model is given the recorded action sequence, and its predicted latents are compared with the latents inferred from the corresponding observed trajectory. At each horizon, we compute $R^2$ across episodes separately for each latent coordinate and aggregate coordinates using target-variance weighting. Scores are computed separately for each environment and model seed, and each model condition is summarized by its mean across four seeds (Appendix~\ref{app:lrc-computation}). Because LRC is computed in each model's own latent space, we validate it against physical rollout consistency (PRC), which decodes the predicted latents with the probes of Section~\ref{sec:eval-probing} and computes $R^2$ against the true states. LRC and PRC are strongly correlated across models and environments (Spearman $\rho=0.81$; Appendix~\ref{app:lrc-validation}). As a state-space topline, we train a two-hidden-layer MLP to predict $\vs_{t+1}$ from $(\vs_t,\va_t)$ and roll it out autoregressively. We also report persistence ($\hat{\vs}_{t+H} = \vs_0$) as a baseline. For both state-space baselines, we set $\vz_t=\vs_t$ and compute LRC directly between the predicted and ground-truth state trajectories.

\paragraph{State decoding (probing).}
\label{sec:eval-probing}
To measure which physical quantities are encoded in each model's latent representation, we freeze the world model and train a separate nonlinear probe to predict each state coordinate $s_t^{(j)}$ from $\vz_t$. Each probe is a two-hidden-layer MLP with widths 128 and 64. We split complete episodes into 80\% for training and 20\%. On the held-out episodes, we report MAE in physical units and as a percentage of the MAE of a shared shuffled-target control. Appendix~\ref{app:probing} provides full details.

\paragraph{Reward evaluation.}
\label{sec:eval-reward}
We report the sum of the \texttt{dm\_control} rewards over 500-step episodes, using the learned controller for TD-MPC2 and DreamerV3 and post-hoc MPC for LeWM. Appendix~\ref{app:reward-eval} provides the controller and evaluation details.

\subsection{Controlled interventions}

To isolate specific failure modes, we run two follow-up experiments on the CartPole environment.
\textit{Nonlinear mixing}: ground-truth state is passed through 0--3 fixed, invertible mixing layers before encoding; the simulator dynamics are unchanged.
\textit{Velocity ablation}: separate pixel models are trained with ground-truth velocities fused into the visual representation ($+v$); as a control, temporal context is reduced from three frames to one for TD-MPC2 and LeWM. Implementation details for both interventions are provided in Appendix~\ref{app:interventions}.

\begin{figure}[t]
  \centering
  \includegraphics[width=\linewidth]{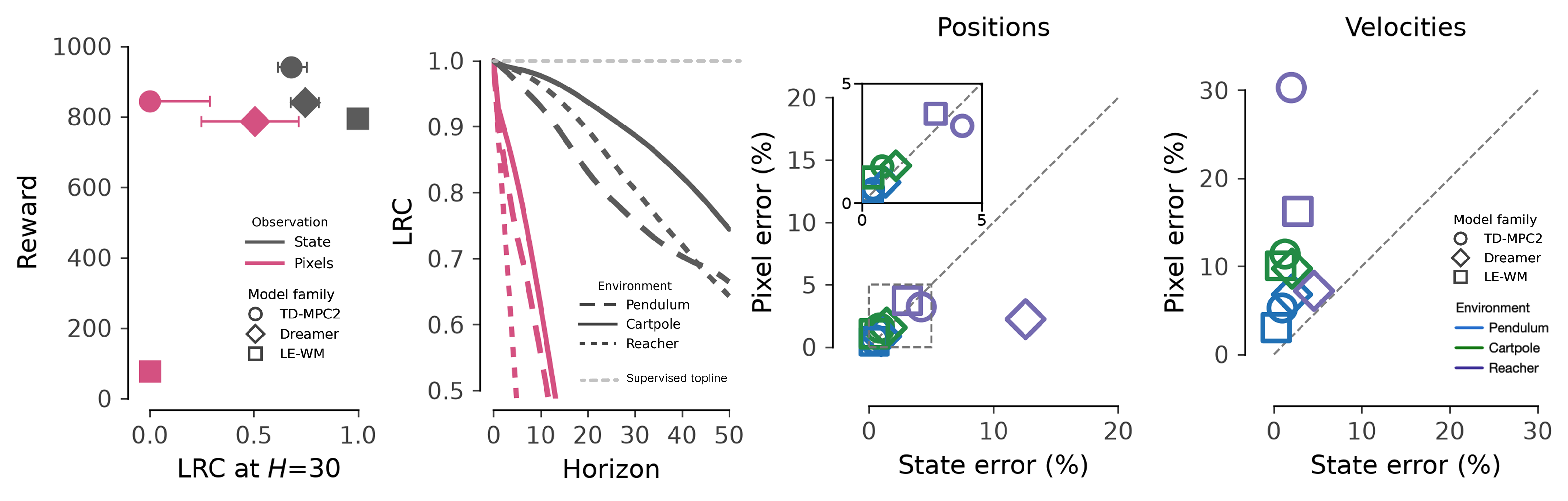}
  \caption{
    \textbf{Latent rollout consistency (LRC) vs. Reward.}
    (a) LRC vs. reward averaged across environments. The 95\% CI is estimated by bootstrap over $n=4$ model seeds.
    (b) LRC over rollout horizons. %
    (c) Position-decoding errors for state- and pixel-based models across environments, normalized by a shuffled-target baseline.
    (d) Velocity-decoding errors, shown as in (c).
  }
    \label{fig:lrc-reward-consistency}
\end{figure}

\section{Results}
\label{sec:results}

\subsection{Task performance does not imply accurate latent prediction}

Task reward is the primary evaluation metric for current latent world models. We compare episode reward with latent rollout consistency (LRC, see Eq. \ref{eq:lrc}), which measures how closely an open-loop latent rollout follows the representations inferred from the corresponding observed trajectory.

\begin{wrapfigure}[19]{r}{0.38\linewidth}
  \centering
  \includegraphics[width=\linewidth]{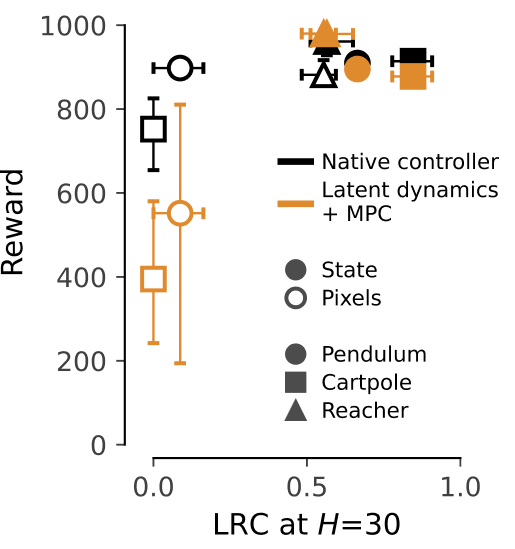}
  \caption{\textbf{Planning with TD-MPC2's frozen latent dynamics alone.} Native controller (black) vs.\ MPC on the latent dynamics (orange).}
  \label{fig:tdmpc2-mpc}
\end{wrapfigure}

Reward and LRC capture distinct aspects of model performance. Models can achieve high rewards while producing inaccurate latent rollouts, even just a few steps into the future (Fig. 4a). This dissociation is most apparent for the task-trained models, TD-MPC2 and DreamerV3, whose controllers can solve the task despite having low LRC. For each model family, we consider the model variant trained on pixels against a topline that receives raw states, including velocities, as the input. Appendix Figure~\ref{fig:reward-lrc-environments} reports the corresponding results separately for each environment.

For task-trained models, this dissociation might arise because their objectives permit task-specific shortcuts. Alternatively, predictive information might be encoded implicitly in policy and value networks, allowing these components to compensate for inaccuracies in the explicit latent dynamics model \citep{letcher2026inverting}.
In contrast, JEPA-based models with a distinct pretraining phase followed by model-predictive control cannot exhibit this shortcut by design. As a consequence, sufficiently high LRC is required to obtain reward, and on the tested environments the quality of the rollouts is not sufficient to meet this bar.
These results show that evaluating latent world models solely through task reward provides an incomplete picture and may obscure whether they learn accurate internal dynamics.

To test this directly, we remove TD-MPC2's reward, value, and policy heads and plan on its frozen latent dynamics with MPC (Fig.~\ref{fig:tdmpc2-mpc}; Appendix~\ref{app:tdmpc2-mpc}). High-LRC models match the native controller, whereas pixel-based Pendulum and CartPole models have near-zero LRC and drop sharply in reward, showing that these task-trained heads compensate for inaccurate dynamics. Without them, swing-up needs longer horizons where pixel rollouts drift, while Reacher is solved in one step (Appendix Fig.~\ref{fig:tdmpc2-mpc-horizon}) and retains higher pixel LRC (Appendix Fig.~\ref{fig:lrc-models-environments}).
The dissociation extends beyond our three environments. For the multi-task TD-MPC2 model, trained from states on 30 DMControl tasks, reward and LRC at $H=30$ are uncorrelated across tasks (Spearman $\rho=-0.11$, $p=0.56$), see Appendix ~\ref{app:multitask}.

\subsection{Pixel observations degrade latent prediction}

\begin{figure}[t]
  \centering
  \includegraphics[width=\linewidth]{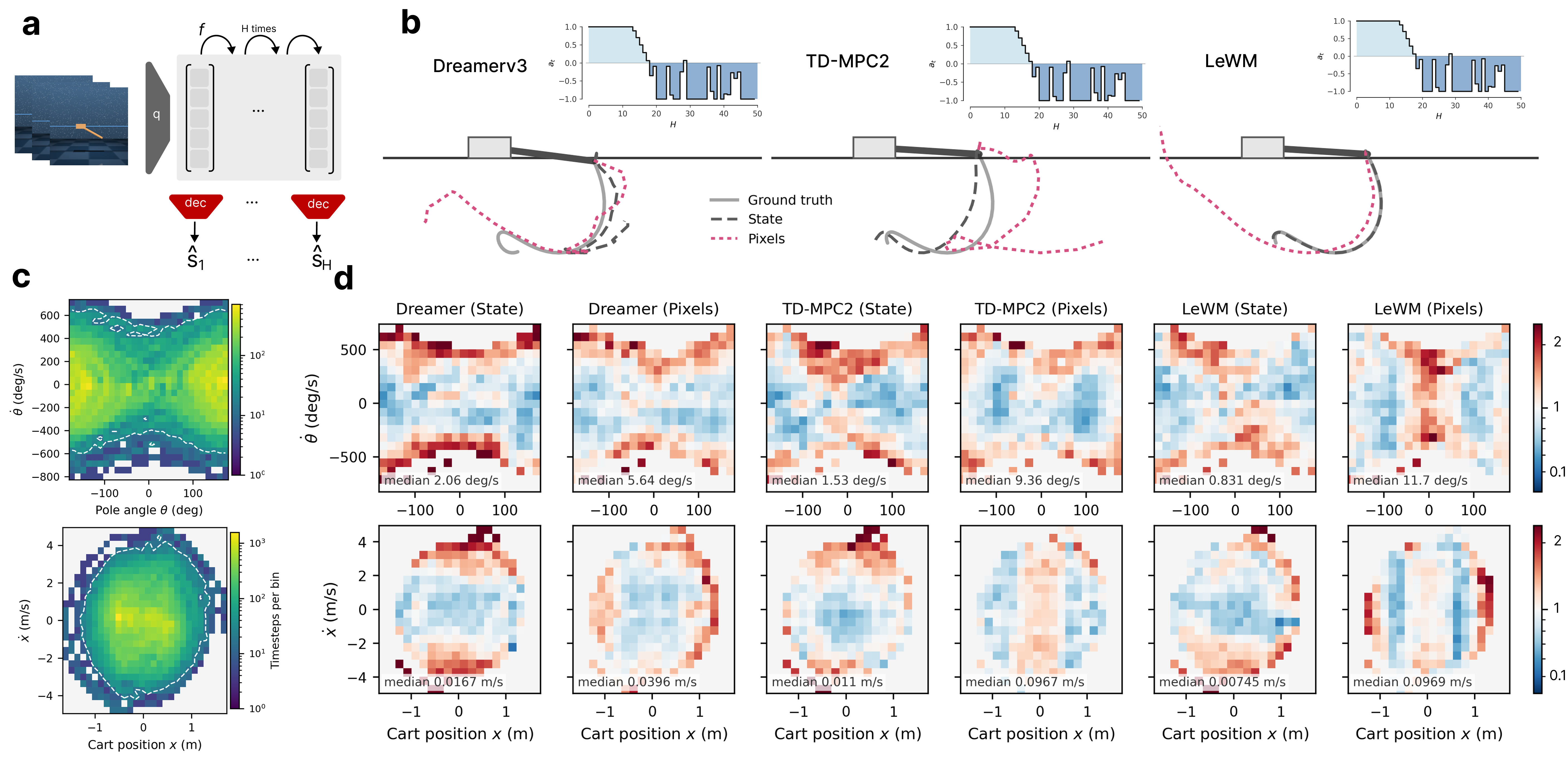}
  \caption{\textbf{CartPole rollouts and state-dependent prediction errors.}
  (a) Predicted latent rollouts are mapped to physical state using the frozen nonlinear probes.
  (b) Example pole-angle trajectories for DreamerV3, TD-MPC2, and LeWM, with the recorded actions shown above each trajectory.
  (c) Evaluation-set occupancy over pole angle and angular velocity (top) and cart position and velocity (bottom).
  (d) Prediction errors over the same state spaces for models trained from states or pixels. Color denotes the error within each bin relative to the median prediction error, and the annotation reports the median error in physical units.}
  \label{fig:qualitative-rollouts-decoding-errors}
\end{figure}

Having shown that task reward can obscure poor latent prediction, we next ask where these deficiencies come from. We compare models trained from ground-truth simulator states with models trained from pixels. State inputs directly provide the physical variables needed for prediction. Pixel inputs, by contrast, require the model to infer those variables from a high-dimensional nonlinear observation and from temporal context. This comparison helps to separate limitations in learning the dynamics from limitations in constructing a predictive representation.

In Figure~\ref{fig:lrc-reward-consistency}b, we report LRC over increasing rollout horizons, averaged across models for each environment. We include a supervised state-space MLP as a topline and persistence as a naive baseline (Section~\ref{sec:eval-lrc}). Although prediction quality degrades in both modalities, pixel-based rollouts deteriorate faster and approach the persistence baseline. The same pixel--state gap is visible within individual model families (Appendix Figure~\ref{fig:lrc-models-environments}).
Figure~\ref{fig:qualitative-rollouts-decoding-errors}b illustrates this behavior through CartPole rollouts decoded using the nonlinear probes described in Section~\ref{sec:eval-probing}. Across all three model families, predictions from pixels depart from the realized trajectory even over the relatively short horizon of 30 control steps, corresponding to only 0.6 seconds of simulated time ($\Delta t=0.02$ s). 

Accurate latent prediction requires both an encoder that retains the predictive state information and a latent dynamics model that evolves that information correctly (Section~\ref{sec:theory}, Eq.~\ref{eqn:faithfulwm}). Low LRC can therefore arise from a failure of either component, or both.
We first examine what the encoder captures by decoding individual state coordinates from frozen latents using nonlinear probes to compare normalized decoding errors for positions (linear positions and angles, Fig.~\ref{fig:lrc-reward-consistency}c) and velocities (linear and angular velocities, Fig.~\ref{fig:lrc-reward-consistency}d). Positions are decoded with low error (0--7\%) in most conditions, with relatively small differences between state and pixel inputs. Velocities are decoded less accurately (3--22\%), and state models consistently outperform pixel models by a large margin. Appendix Figures~\ref{fig:state-decoding-physical} and~\ref{fig:state-decoding-probes} report the per-environment errors in physical units and compare linear with nonlinear probes, respectively.

Figure~\ref{fig:qualitative-rollouts-decoding-errors}d shows angular-velocity errors (top row) and cart-velocity errors (bottom row) across the CartPole state space. These errors are not uniformly distributed: they increase with large velocities, where the system evolves most rapidly and estimation errors can compound over autoregressive rollouts. This pattern appears for both input modalities, but the median error is roughly $2$--$15\times$ larger for pixel-based models, depending on the model and velocity variable.

\subsection{Missing velocities explain part of the pixel--state gap}
\label{sec:velocity-intervention}

\begin{figure}[t]
  \centering
  \includegraphics[width=1.\linewidth]{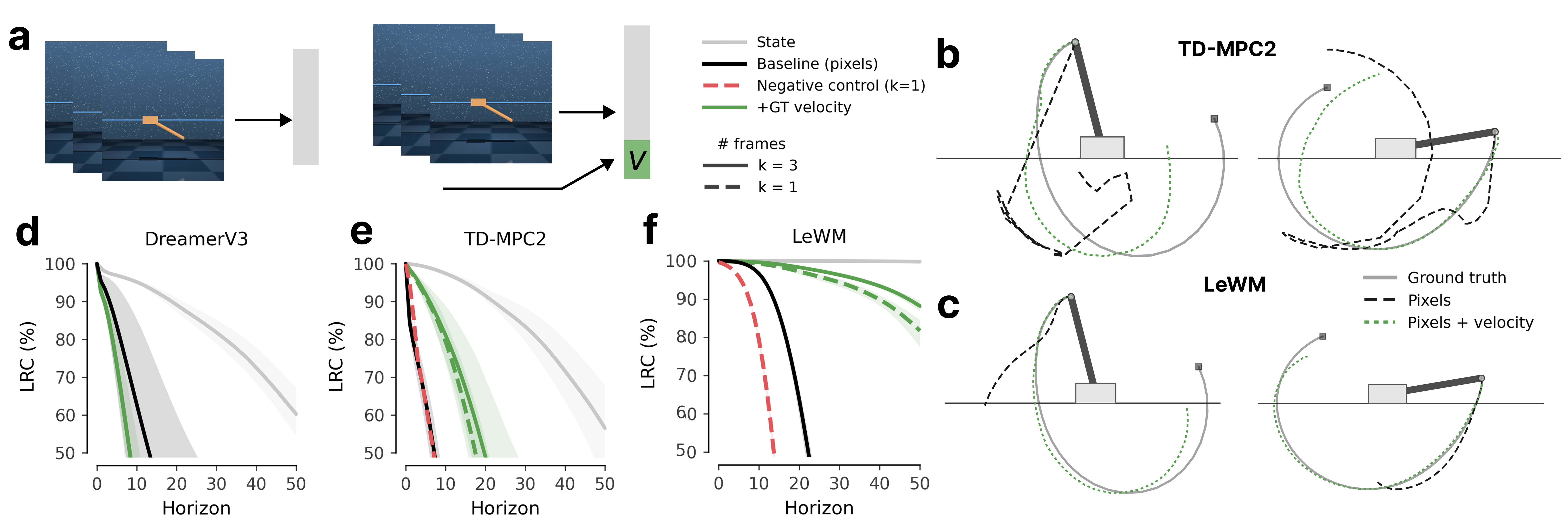}
  \caption{\textbf{Velocity observability and world-model prediction.} (a) Ground-truth velocity is fused with the visual representation. (b--c) Example CartPole rollouts with and without velocity for TD-MPC2 and LeWM. (d--f) LRC over the rollout horizon for DreamerV3, TD-MPC2, and LeWM. Full results are reported in Table~\ref{tab:velocity-exp}.}
  \label{fig:velocity-observability}
\end{figure}

This asymmetry likely arises because the underlying physical state is Markovian, whereas a single pixel observation is not and velocities can only be inferred from temporal context. We test this hypothesis on CartPole by providing pixel-based models with ground-truth cart and pole velocities and comparing LRC, task reward, and velocity decoding error (Figure~\ref{fig:velocity-observability}). During evaluation, velocity is provided only at initialization; the latent dynamics must evolve the state thereafter. As a negative control, we also reduce the temporal context of TD-MPC2 and LeWM from three frames to a single frame. Full implementation details are provided in Appendix~\ref{app:velocity-exp}.

\begin{table}[t]
  \centering
  \footnotesize
  \setlength{\tabcolsep}{2.5pt}
  \begin{subtable}[t]{0.30\textwidth}
    \centering
    \caption{DreamerV3}
    \label{tab:dreamer-compact}
    \resizebox{\linewidth}{!}{%
    \begin{tabular}{@{}lrrrr@{}}
    \toprule
    Input & LRC$_{10}$ & LRC$_{30}$ & R & $\epsilon_{\dot q}$ \\
    \midrule
    state & \textbf{95.2} & 81.9 & \textbf{871} & 2.28 \\
    \midrule
    frames & 63.8 & 8.9 & \textbf{872} & 9.51 \\
    $\quad$+v & 37.9 & < 0 & \textbf{774} & 8.06 \\
    \bottomrule
    \end{tabular}}
  \end{subtable}%
  \hspace{1.2em}%
  \begin{subtable}[t]{0.30\textwidth}
    \centering
    \caption{TD-MPC2}
    \label{tab:tdmpc2-compact}
    \resizebox{\linewidth}{!}{%
    \begin{tabular}{@{}lrrrr@{}}
    \toprule
    Input & LRC$_{10}$ & LRC$_{30}$ & R & $\epsilon_{\dot q}$ \\
    \midrule
    state & \textbf{97.4} & 83.4 & \textbf{878} & 1.18 \\
    \midrule
    3 frames & 29.6 & < 0 & \textbf{832} & 11.45 \\
    $\quad$+v & 80.8 & 16.4 & \textbf{855} & 1.90 \\
    1 frame & 27.7 & < 0 & 206 & 130.61 \\
    $\quad$+v & 79.5 & 5.0 & \textbf{865} & 1.81 \\
    \bottomrule
    \end{tabular}}
  \end{subtable}%
  \hspace{1.2em}%
  \begin{subtable}[t]{0.30\textwidth}
    \centering
    \caption{LeWM}
    \label{tab:lewm-compact}
    \resizebox{\linewidth}{!}{%
    \begin{tabular}{@{}lrrrr@{}}
    \toprule
    Input & LRC$_{10}$ & LRC$_{30}$ & R & $\epsilon_{\dot q}$ \\
    \midrule
    state & \textbf{100.0} & \textbf{99.9} & \textbf{757} & 0.73 \\
    \midrule
    3 frames & \textbf{96.7} & 1.2 & 151 & 10.16 \\
    $\quad$+v & \textbf{99.6} & 96.3 & 178 & 1.29 \\
    1 frame & 79.4 & < 0 & 171 & 130.39 \\
    $\quad$+v & \textbf{99.3} & 94.4 & 195 & 1.28 \\
    \bottomrule
    \end{tabular}}
  \end{subtable}
  \caption{LRC at $H=10$ and $H=30$, reward ($R$), and normalized velocity error ($\epsilon_{\dot q}$); $+v$ add ground-truth velocity. LRC and errors are in percentage units. Bold: $R>700$ or LRC$_{10}>90$.}
  \label{tab:velocity-exp}
\end{table}

Adding velocity substantially improves TD-MPC2 and LeWM, both with reduced and full temporal context. However, it has no effect on DreamerV3. This could be partly because the velocity is available to its RSSM posterior but not its prior, potentially creating a mismatch.
The decoding results (Table \ref{tab:velocity-exp}, $\epsilon_{\dot{q}}$) confirm that the intervention has the intended effect on the representation. Without velocity injection, reducing the temporal context to one frame raises normalized velocity error to approximately 130\% for both TD-MPC2 and LeWM. Providing velocity reduces this error to approximately 1--2\% (Table~\ref{tab:velocity-exp}). 
Overall, while adding velocity explains part of the LRC gap between pixel and state inputs, it does not fully account for the remaining difference in rollout accuracy.

Improved velocity representation does not consistently increase task reward. TD-MPC2 changes little, LeWM remains far below its state-input performance despite substantially improving LRC, and DreamerV3 performs worse. CartPole swing-up requires coordinated actions over a longer timescale than 30 steps, which corresponds to only 0.3 seconds. LeWM’s improved LRC at $H=30$ is therefore insufficient for planning the complete swing-up motion, consistent with its limited reward improvement.

\subsection{Learning latent dynamics from nonlinear observations}
\label{sec:nonlinear-mixing}

\begin{figure}[t]
  \centering
  \includegraphics[width=1.\linewidth]{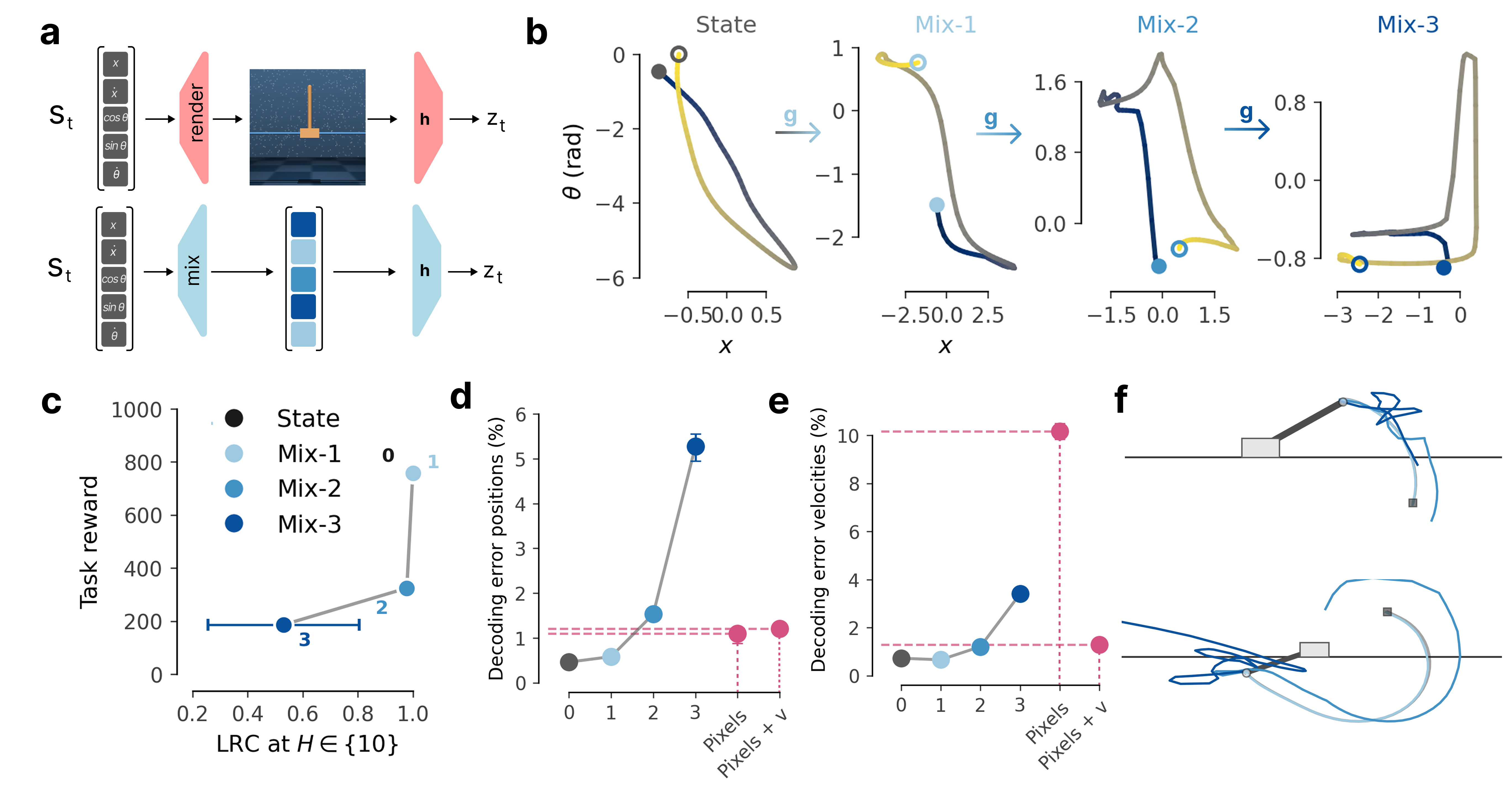}
  \caption{\textbf{World modeling under nonlinear observation mixing on CartPole.} (a--b) Pixel rendering is contrasted with approximately invertible mixing of the underlying state. (c--f) LRC at $H=10$, task reward, normalized position error, and normalized velocity error for LeWM versus mixing depth, with pixels shown as a reference.}
  \label{fig:nonlinear-observation-mixing}
\end{figure}

Missing velocity explains only part of the pixel--state gap. Another key difference is that \textit{pixels are a nonlinear transformation of the underlying physical state}. On CartPole, we replace rendering with an invertible nonlinear observation map whose complexity we can control. We then ask whether models can still learn faithful latent dynamics when all state information is preserved but represented in increasingly nonlinear coordinates. 
We present results for LeWM in Figure~\ref{fig:nonlinear-observation-mixing} because its self-supervised objective most directly approximates the faithful condition in Eq.~\ref{eqn:faithfulwm}. This allows us to study the effect of nonlinear observations with fewer task-specific confounds. Appendix Figure~\ref{fig:nonlinear-mixing-all-models} reports the same intervention for TD-MPC2 and DreamerV3 alongside LeWM.
For each world model, we pass the ground-truth state vector through a bijective mixing function~\citep{hyvarinen2019nonlinear} with 0--3 layers before the encoder (Figure~\ref{fig:nonlinear-observation-mixing}a--b). Because the mixing function is invertible, \textit{the mixed state vector contains the same information as the raw state, only in a harder-to-invert coordinate system}. We include the standard pixel condition as a reference point on the same plots.

Task reward and LRC jointly decline as the number of mixing layers increases (Figure~\ref{fig:nonlinear-observation-mixing}c), and the controller solves only the zero-layer (raw-state) and one-layer conditions.
We next examine representation quality in these conditions. The pixel model has a normalized position-decoding error slightly above 1\% and does not solve the task. With one mixing layer, the error is approximately 0.5\% and the model solves the task; with two layers, the error exceeds 1.5\% and the task fails (Figure~\ref{fig:nonlinear-observation-mixing}c--d).
Velocity encoding may also contribute to this behavior. The pixel-plus-velocity condition from Figure~\ref{fig:velocity-observability} has a velocity-decoding error comparable to that obtained with two mixing layers (Figure~\ref{fig:nonlinear-observation-mixing}e).
Qualitative rollouts show that errors arise early and severely degrade rollout quality for models with more than one mixing layer.
These failures cannot be attributed to missing information in the model input. %

\section{Discussion and Limitations}
\label{sec:discussion}

Our experiments show that task performance alone provides limited information about the dynamics learned by a latent world model. Using LRC and state decoding, we point to two sources of difficulty in learning faithful dynamics. First, pixel-based models struggle to recover velocity. Providing velocity explicitly improves latent prediction but does not fully close the gap to state-based models. Second, even when the underlying state is fully observed, nonlinear transformations make both the representation and its dynamics substantially harder to learn. 

Our work deliberately focuses on simple deterministic continuous-control environments. This makes the physical state and transition dynamics observable and allows direct evaluation of faithful latent dynamics, but leaves open how these results extend to stochastic, partially observable, discrete, or substantially higher-dimensional environments. Similarly, we evaluate three representative latent world-model families rather than attempting an exhaustive comparison of existing approaches. \textit{Our results should be interpreted as controlled counterexamples to the assumption that successful control implies an accurate model of the underlying environment dynamics, rather than as evidence that all latent world models exhibit the same failure modes}.

\section{Conclusion}

We asked whether latent world models that successfully solve control tasks actually learn representations and dynamics that model their environment. To make this question testable, we defined \emph{faithful latent world modeling} through an equivariance condition and introduced latent rollout consistency (LRC) together with state decoding to measure these properties independently of task reward. Across three latent world models and controlled continuous-control environments, high reward did not imply faithful latent dynamics, particularly when learning from pixels.

Our results further show that this gap cannot be attributed to a single source of failure: recovering the dynamical state from observations is difficult, but even with the relevant state information available, nonlinear observations can make the dynamics substantially harder to learn. These results suggest that improving world models requires addressing representation and dynamics learning jointly, and motivate closing the gap between theory of identifiable representation learning and numerically successful implementations of these algorithms.

\section*{AI use statement}
We used generative AI tools to assist with drafting and editing the manuscript, implementing and running experimental code, and creating plotting code for the figures. All AI-assisted text and code were reviewed by the authors. Experimental outputs and figures were checked for correctness, and the authors take full responsibility for the final content of this work.

\ifanon
\bibliographystyle{iclr2027_conference}
\else
\bibliographystyle{unsrtnat}
\fi
\bibliography{references}

\appendix

\clearpage
\etocdepthtag.toc{appendix}
\etocsettagdepth{main}{none}
\etocsettagdepth{appendix}{subsection}
\renewcommand{\contentsname}{Appendix contents}
\tableofcontents
\let\appendixsection\section
\renewcommand{\section}{\clearpage\appendixsection}

\section{Environments and evaluation datasets}
\label{app:envs_data}

\paragraph{Environments.}
We evaluate on three DeepMind Control Suite \citep{tunyasuvunakool2020} environments: \textit{pendulum}, \textit{cartpole}, and \textit{reacher}.
Each exposes a low-dimensional simulator state $\vs_t$ that we use as the probing target, and image observations rendered at $64{\times}64$ and $224{\times}224$ for pixel-based world models.
Pendulum (pendulum-swingup) has a 1-dimensional torque action and a 3-dimensional state $(\cos\theta, \sin\theta, \dot{\theta})$.
CartPole (cartpole-swingup) has a 1-dimensional force action and a 5-dimensional state $(x, \cos\theta, \sin\theta, \dot{x}, \dot{\theta})$ following the standard DM-Control parameterization.
Reacher (reacher-hard) has a 2-dimensional torque action and a 6-dimensional state consisting of two joint angles, two joint velocities, and the 2D Cartesian target offset.
Table~\ref{tab:environments} summarizes all three environments.

\begin{table}[H]
  \centering
  \caption{Evaluation environments and the underlying physical state used as probing targets.}
  \label{tab:environments}
  \small
  \setlength{\tabcolsep}{4pt}
  \begin{tabularx}{\linewidth}{@{}l c c >{\raggedright\arraybackslash}X >{\raggedright\arraybackslash}X@{}}
    \toprule
    \textbf{Env.} & \textbf{$d_s$} & \textbf{$d_a$} & \textbf{State variables} & \textbf{Task} \\
    \midrule
    Pendulum (pendulum-swingup) &
    3 & 1 &
    $\cos\theta$, $\sin\theta$, pole angular velocity &
    Swing the pole upright from below and balance it. \\
    \addlinespace
    CartPole (cartpole-swingup) &
    5 & 1 &
    cart position, $\cos$/$\sin$ of pole angle, cart and pole velocities &
    Swing the pole upright from below and balance it by moving the cart. \\
    \addlinespace
    Reacher (reacher-hard) &
    6 & 2 &
    two joint angles, two joint velocities, 2D target offset &
    Bring the fingertip of a two-link arm to a small randomized target. \\
    \bottomrule
  \end{tabularx}
\end{table}

\paragraph{Evaluation data.} For each environment, we collect 1,000 episodes of 100 steps each. Each episode stores the simulator state $\vs_t$, the action $\va_t$, the reward, and rendered $64{\times}64$ and $224{\times}224$ RGB frames.

\begin{figure}[H]
  \centering
  \includegraphics[width=0.9\linewidth]{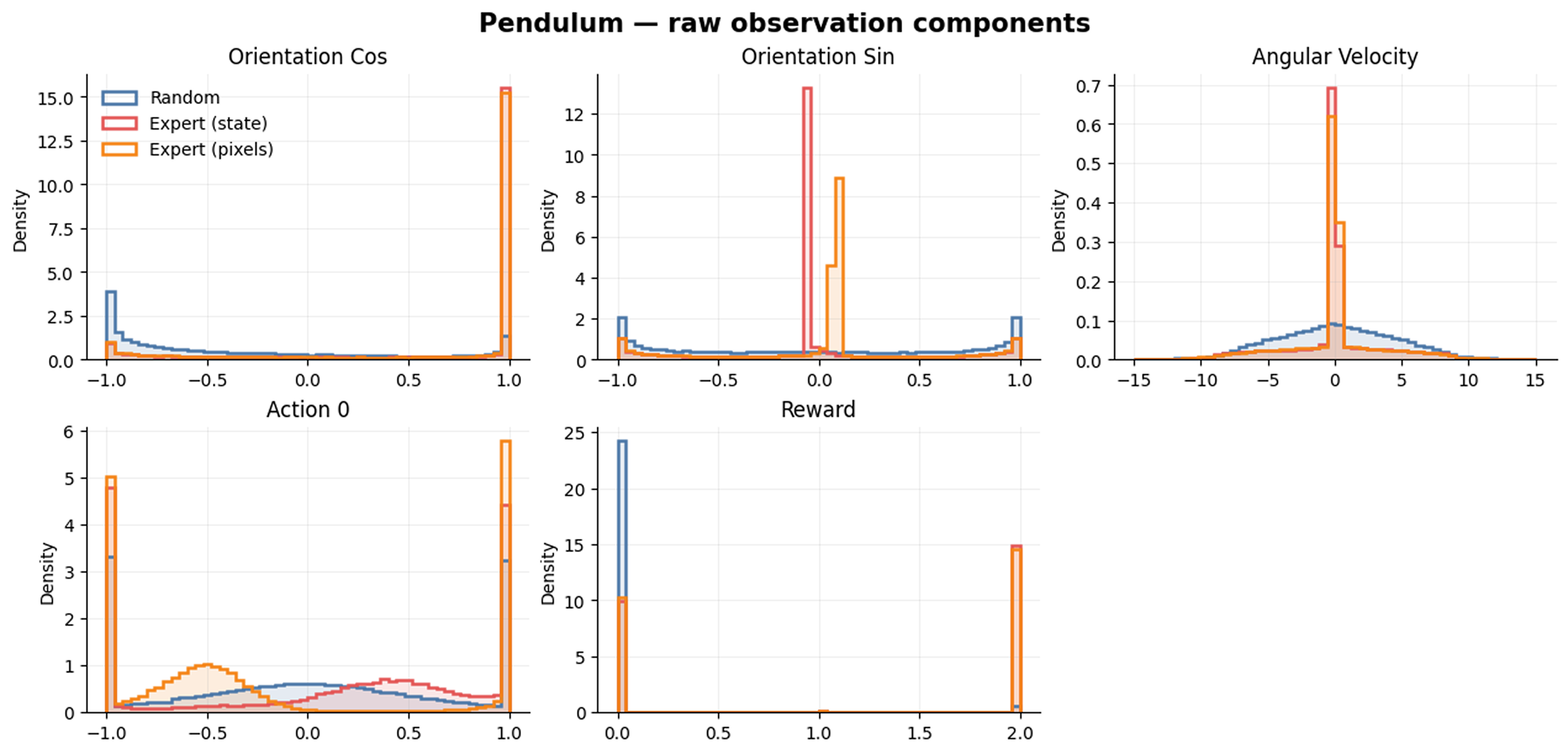}
  \caption{\textbf{Pendulum data distributions.}
  Marginal distributions of the orientation coordinates, angular velocity, action, and native \texttt{dm\_control} reward in the random OU and expert datasets.}
  \label{fig:data-distribution-pendulum}
\end{figure}

\begin{figure}[H]
  \centering
  \includegraphics[width=\linewidth]{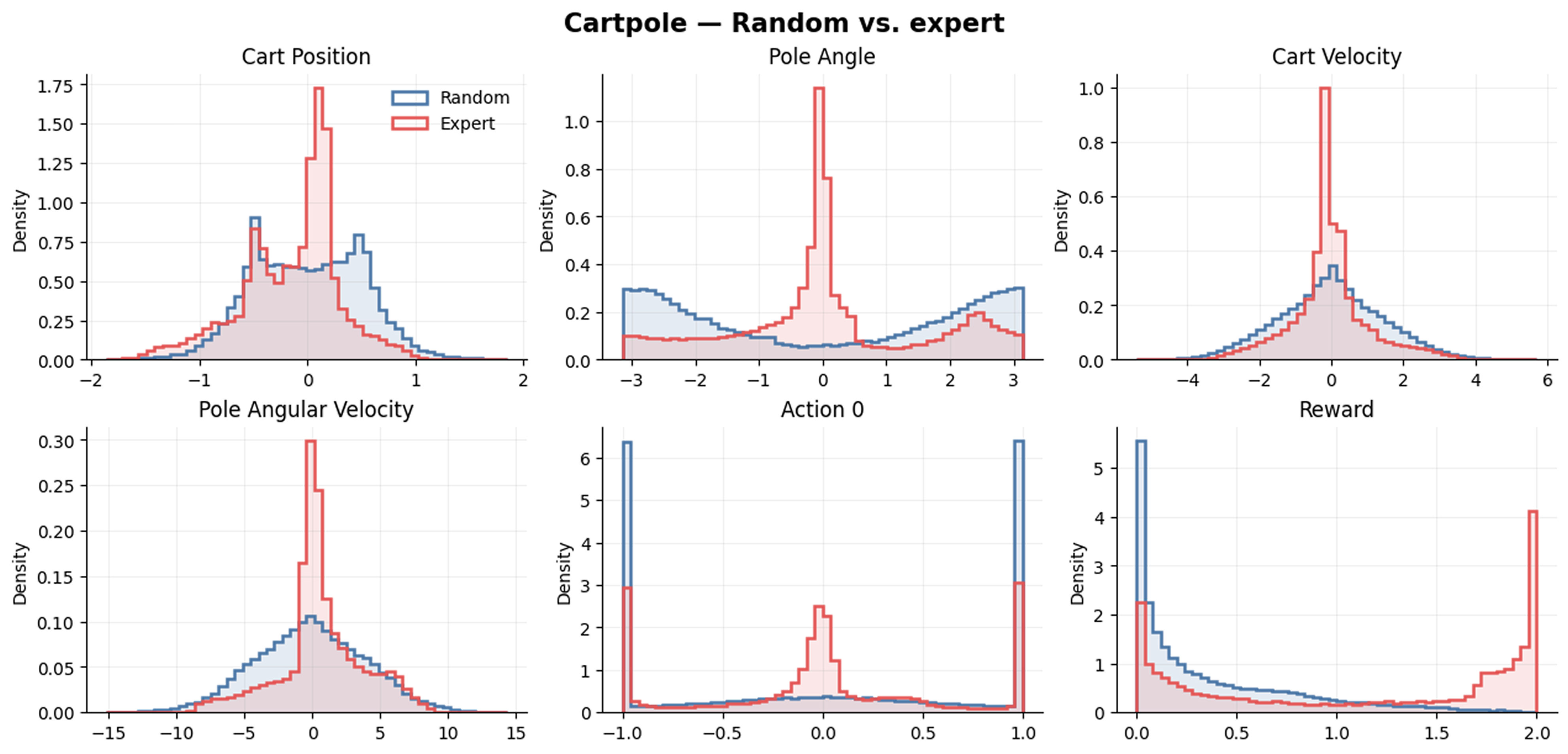}
  \caption{\textbf{CartPole data distributions.}
  Marginal distributions of cart position, pole angle, cart and pole velocities, action, and native \texttt{dm\_control} reward in the random OU and expert datasets.}
  \label{fig:data-distribution-cartpole}
\end{figure}

\begin{figure}[H]
  \centering
  \includegraphics[width=\linewidth]{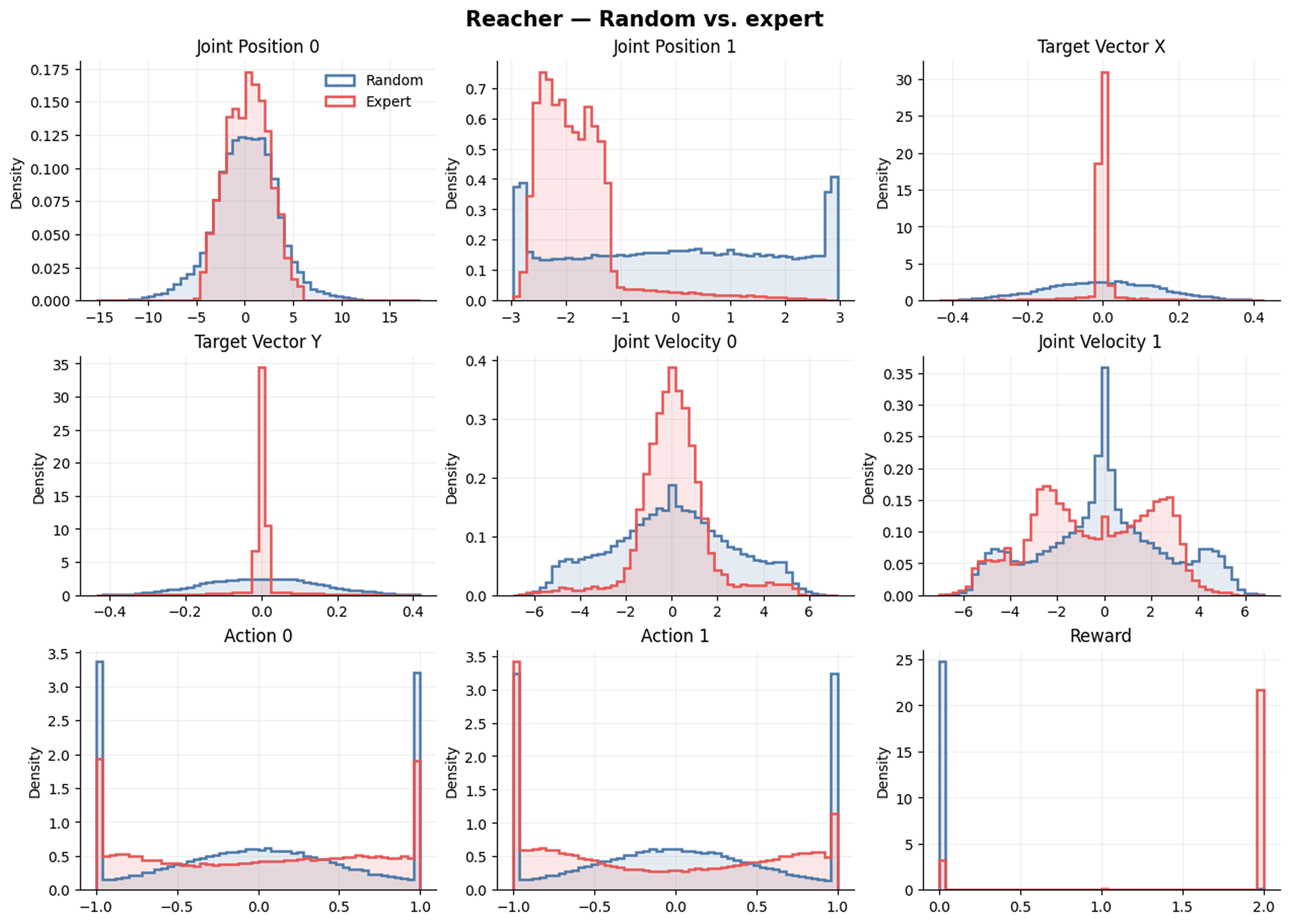}
  \caption{\textbf{Reacher data distributions.}
  Marginal distributions of the joint positions, target-offset coordinates, joint velocities, two action coordinates, and native \texttt{dm\_control} reward in the random OU and expert datasets.}
  \label{fig:data-distribution-reacher}
\end{figure}

Actions are drawn from an Ornstein--Uhlenbeck (OU) process with mean reversion $\theta=0.15$ and base diffusion $\sigma=0.35$, modulated by intermittent high-energy bursts to encourage broad coverage of joint angles and velocities.
The OU schedule is identical across environments except that, for CartPole, a higher fraction of episodes uses the high-energy regime ($0.75$ vs.\ $0.4$ for Pendulum and Reacher) and the cart position is softly constrained to $[-0.85, 0.85]$ via a position-keeper that becomes active near the rails.
Initial states are sampled from broad priors over joint positions and velocities (uniform over angles, near-zero velocities), so that the resulting trajectories visit a wide range of physical states without using a trained controller. Figures~\ref{fig:data-distribution-pendulum}--\ref{fig:data-distribution-reacher} compare the marginal distributions of every state variable, action, and native \texttt{dm\_control} reward under the random OU and expert policies.

\section{Latent rollout consistency (LRC)}
\label{app:lrc-computation}

\subsection{Computation}
\label{app:lrc-definition}

We initialize one autoregressive latent rollout at the beginning of every evaluation episode $e$. Starting from the encoded initial observation, $\hat{\vz}_{e,0}=\vz_{e,0}$, the learned dynamics are applied to the recorded actions, $\hat{\vz}_{e,t+1}=\hat{\rvf}(\hat{\vz}_{e,t},\va_{e,t})$, and the prediction at horizon $H$ is compared with the target $\vz_{e,H}=\rvh(\vx_{e,\leq H},\va_{e,<H})$ inferred from the realized observations. Stacking the targets $\vz_{e,H}$ and predictions $\hat{\vz}_{e,H}$ of the $N$ evaluation episodes into $\mZ_H,\hat{\mZ}_H\in\mathbb{R}^{N\times d_z}$, latent rollout consistency is
\begin{equation}
  \mathrm{LRC}(H)=R^2\bigl(\mZ_H,\hat{\mZ}_H\bigr).
  \label{eq:lrc-matrix}
\end{equation}
We use the variance-weighted $R^2$, which averages the per-coordinate $R^2$ values across episodes with weights proportional to each coordinate's target variance. $\mathrm{LRC}(H)=1$ indicates perfect prediction and $\mathrm{LRC}(H)=0$ indicates predicting the mean.

\subsection{Validation}
\label{app:lrc-validation}

LRC is computed in each model's own latent space, and these spaces differ in dimensionality, scale and structure. We test whether LRC measures prediction quality consistently across models by comparing it with a reference metric computed in a shared physical space. We then compare LRC with alternative latent-space measures of prediction error (cosine similarity, $L_2$ distance) by how closely each agrees with this reference.

\paragraph{Physical rollout consistency.}
Physical rollout consistency (PRC) decodes each predicted latent with the frozen nonlinear probes $\rvp:\gZ\to\gS$ of Appendix~\ref{app:probing} and compares the result with the true simulator states $\vs_{e,H}$, stacked into $\mS_H\in\mathbb{R}^{N\times d_s}$,
\begin{equation}
  \mathrm{PRC}(H) = R^2\bigl(\mS_H, \rvp(\hat{\mZ}_H)\bigr),
\end{equation}
where per-coordinate $R^2$ values are averaged with equal weights. PRC is computed in the same space for every model and is therefore independent of latent geometry. It requires the ground-truth state and a trained probe.

\begin{figure}[H]
  \centering
  \includegraphics[width=0.9\linewidth]{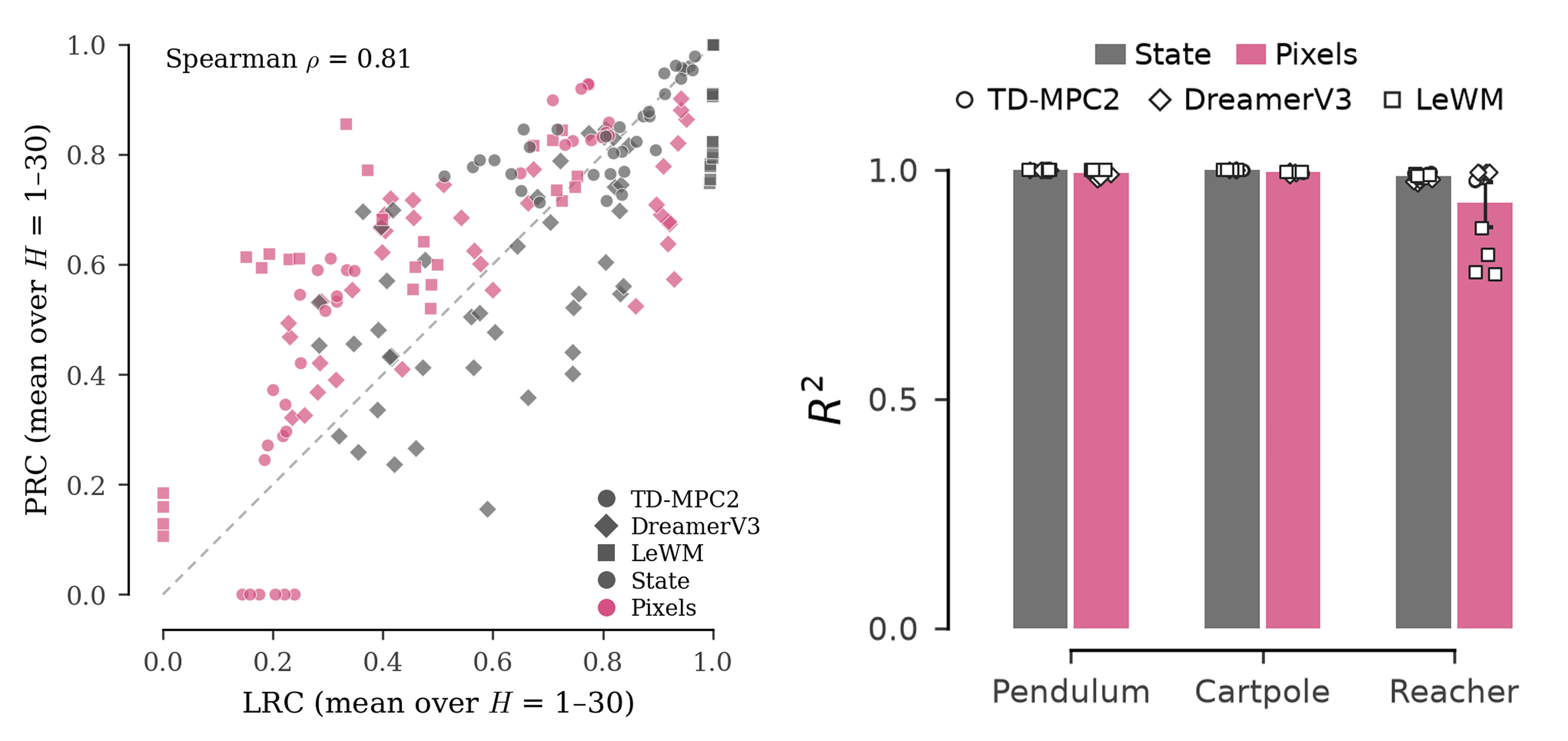}
  \caption{\textbf{LRC agrees with prediction accuracy in physical space.}
  Left: LRC against PRC, averaged over horizons $H=1$--$30$. Right: probe decoding $R^2$ at $H=0$ for each environment and input modality.}
  \label{fig:lrc-vs-prc}
\end{figure}

\paragraph{LRC tracks prediction accuracy in physical space.}
Averaged over horizons $H=1$--$30$, LRC is strongly correlated with PRC across models, modalities and environments (Figure~\ref{fig:lrc-vs-prc}, left; $\rho = 0.81$). PRC is only reliable if the probes decode the state accurately, which holds at $H=0$, where they reach high $R^2$ (Figure~\ref{fig:lrc-vs-prc}, right). Imagined latents, however, can drift outside the distribution of encoded latents on which the probes were trained, so PRC may also reflect probe errors at longer horizons.

\begin{figure}[H]
  \centering
  \includegraphics[width=\linewidth]{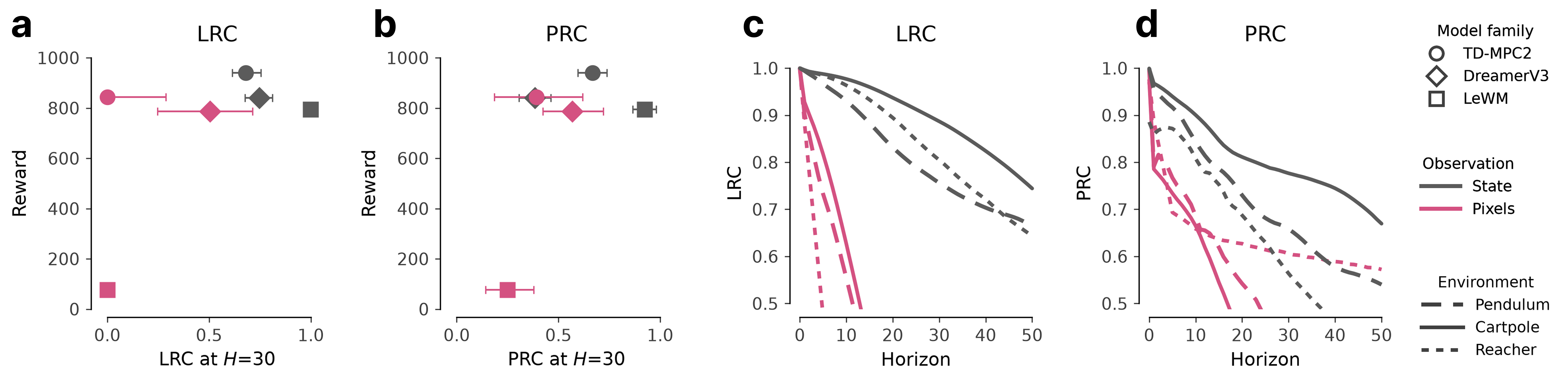}
  \caption{\textbf{Main results evaluated in latent and physical space.}
  (a) LRC and (b) PRC against reward at $H=30$, averaged across environments; (a) corresponds to Figure~\ref{fig:lrc-reward-consistency}a.
  (c) LRC and (d) PRC over rollout horizons, averaged across models for each environment; (c) corresponds to Figure~\ref{fig:lrc-reward-consistency}b.}
  \label{fig:prc-main-results}
\end{figure}

\paragraph{Main results in physical space.}
The main results do not depend on measuring rollouts in latent space. With PRC in place of LRC, high reward still coexists with inaccurate rollouts, and pixel-based models still fall behind state-based ones over the horizon (Figure~\ref{fig:prc-main-results}). PRC is less extreme than LRC for pixel-based models, as most of them lie above the diagonal in Figure~\ref{fig:lrc-vs-prc} (left), but the ordering of models and modalities is unchanged.

\begin{figure}[H]
  \centering
  \includegraphics[width=\linewidth]{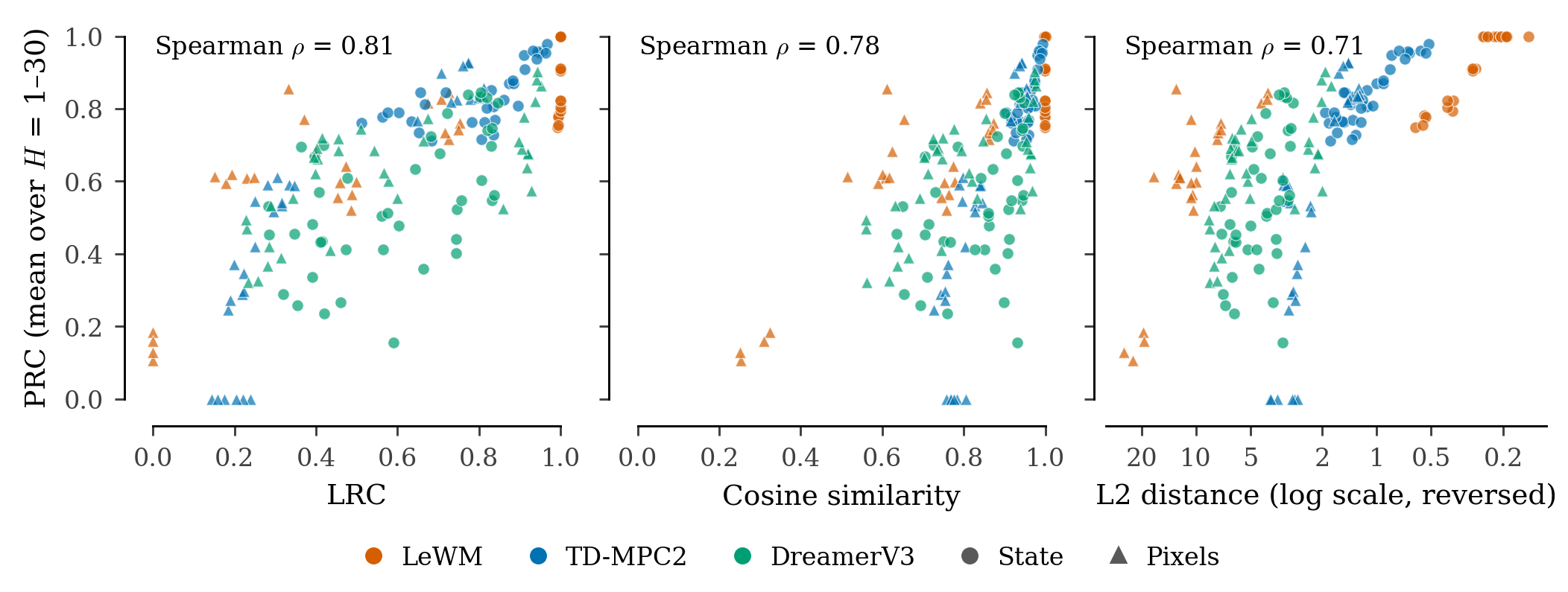}
  \caption{\textbf{Latent-space measures of dynamics equivariance against PRC.} From left to right: LRC, cosine similarity, $L_2$ distance (log scale, reversed). Scores are averaged over horizons $H=1$--$30$. Color indicates the model family, marker shape the input modality.}
  \label{fig:lrc-variants}
\end{figure}

\paragraph{Sensitivity to the choice of metric.}
Figure~\ref{fig:lrc-variants} compares LRC with two alternative latent-space measures by their agreement with PRC, averaged over horizons $H=1$--$30$. LRC agrees most closely ($\rho = 0.81$), followed by cosine similarity ($\rho = 0.78$) and the $L_2$ distance ($\rho = 0.71$). The $L_2$ distance depends on the latent dimensionality $d_z$, and cosine similarity has no fixed reference point, whereas $\mathrm{LRC}=0$ corresponds to predicting the mean.

\subsection{DreamerV3 stochastic rollouts}
\label{app:dreamer-rollouts}

\begin{figure}[H]
  \centering
  \includegraphics[width=\linewidth]{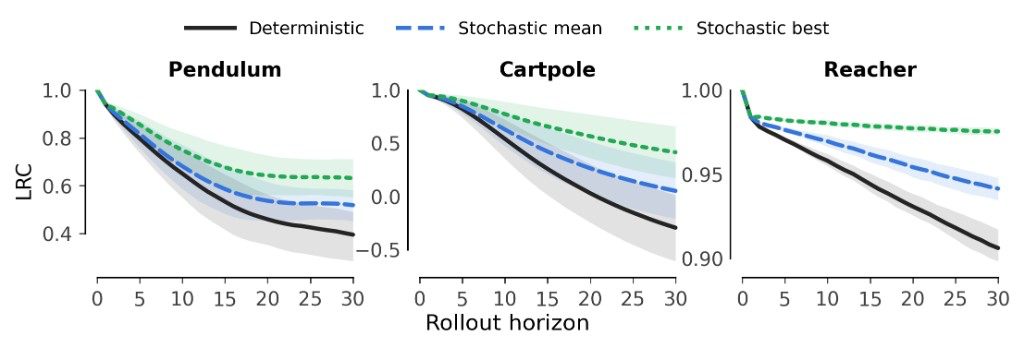}
  \caption{\textbf{Marginalizing DreamerV3's stochastic transitions.}
  LRC over the rollout horizon on Pendulum, CartPole, and Reacher using the deterministic RSSM rollout, the coordinate-wise mean of 500 stochastic rollouts generated from a shared posterior anchor, and the best-matching sampled rollout. The latter uses the realized trajectory to select a sample and therefore represents an oracle upper bound rather than a deployable estimator. Shaded regions show variation across model seeds.}
  \label{fig:dreamer-stochastic-rollouts}
\end{figure}

Because DreamerV3 uses a stochastic RSSM, we infer the filtered posterior once, copy this same anchor across 500 rollouts, replay the recorded actions, and independently sample the categorical prior state at every future step. We represent each predicted state by the deterministic RSSM state concatenated with the categorical probabilities, average these feature vectors coordinate-wise across the 500 trajectories at every horizon, and use this Monte Carlo mean as $\hat{\vz}_{e,H}$ above. At each horizon, we therefore average the 500 predicted latent vectors and compute LRC between this mean prediction and the latent representation inferred from the corresponding observed trajectory. Figure~\ref{fig:dreamer-stochastic-rollouts} compares this estimator with the deterministic RSSM rollout and an oracle that selects the best-matching stochastic rollout from the same samples.

\begin{figure}[H]
  \centering
  \includegraphics[width=\linewidth]{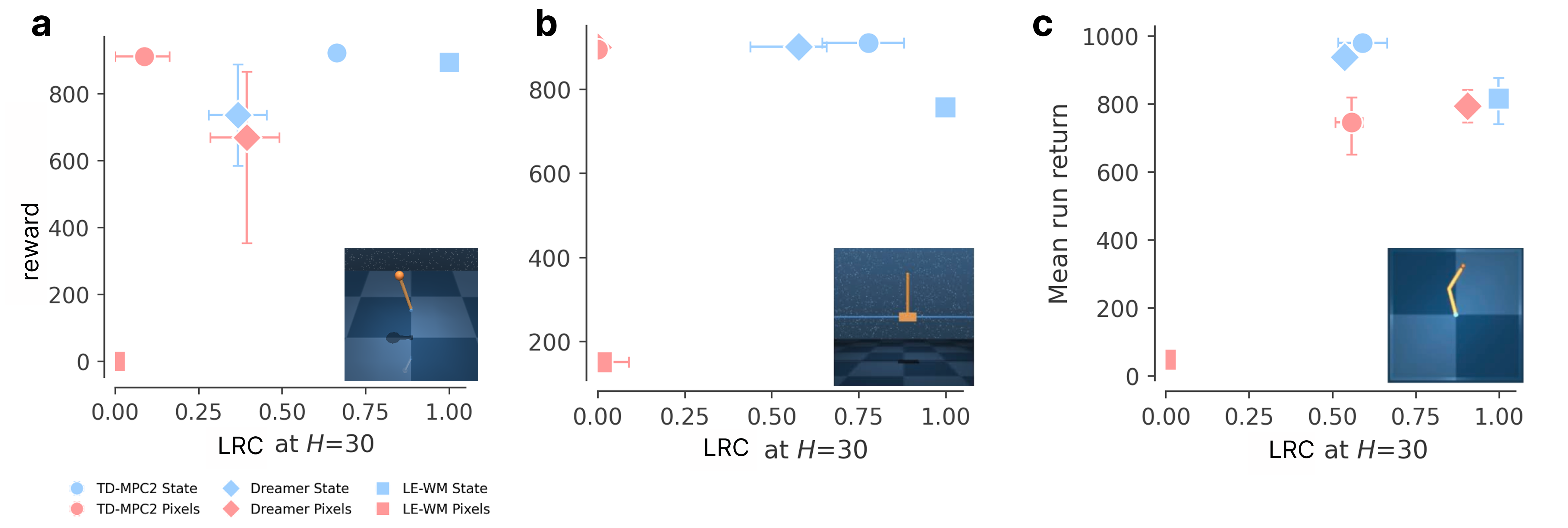}
  \caption{\textbf{Task reward and latent rollout consistency by environment.}
  Episode reward versus LRC at horizon $H=30$ for (a) Pendulum, (b) CartPole, and (c) Reacher. Marker shape identifies the model family and color identifies whether the model receives ground-truth states or pixels; error bars show variation across model seeds.}
  \label{fig:reward-lrc-environments}
\end{figure}

\section{Sensitivity to the data-collection policy}
\label{app:own-policy}

All analyses in the main text are evaluated on trajectories from a random Ornstein--Uhlenbeck (OU) policy, chosen for broad state-space coverage (Appendix~\ref{app:envs_data}). Task reward, however, is measured under each model's own controller, which visits a narrower region of the state space. Low LRC on random-policy data could therefore partly reflect a shift between the states a model is trained and used on and the states it is evaluated on. Here we repeat the main analyses on trajectories generated by each model's own controller and test whether the conclusions change.

\subsection{Own-policy evaluation data}
\label{app:own-policy-data}

For each model seed and environment, we collect an evaluation set with the same number of episodes and steps per episode as the random dataset (Appendix~\ref{app:envs_data}), using the model's own controller to select actions. This is the learned policy for TD-MPC2 and DreamerV3 and post-hoc MPC for LeWM. LRC is then computed on each model's own trajectories exactly as in Appendix~\ref{app:lrc-computation}. Unlike the random dataset, these evaluation sets differ between models, so LRC values on own-policy data compare each model with itself rather than all models on a common test set.

\subsection{Reward and latent rollout consistency on own-policy trajectories}
\label{app:own-policy-main}

\begin{figure}[H]
  \centering
  \includegraphics[width=\linewidth]{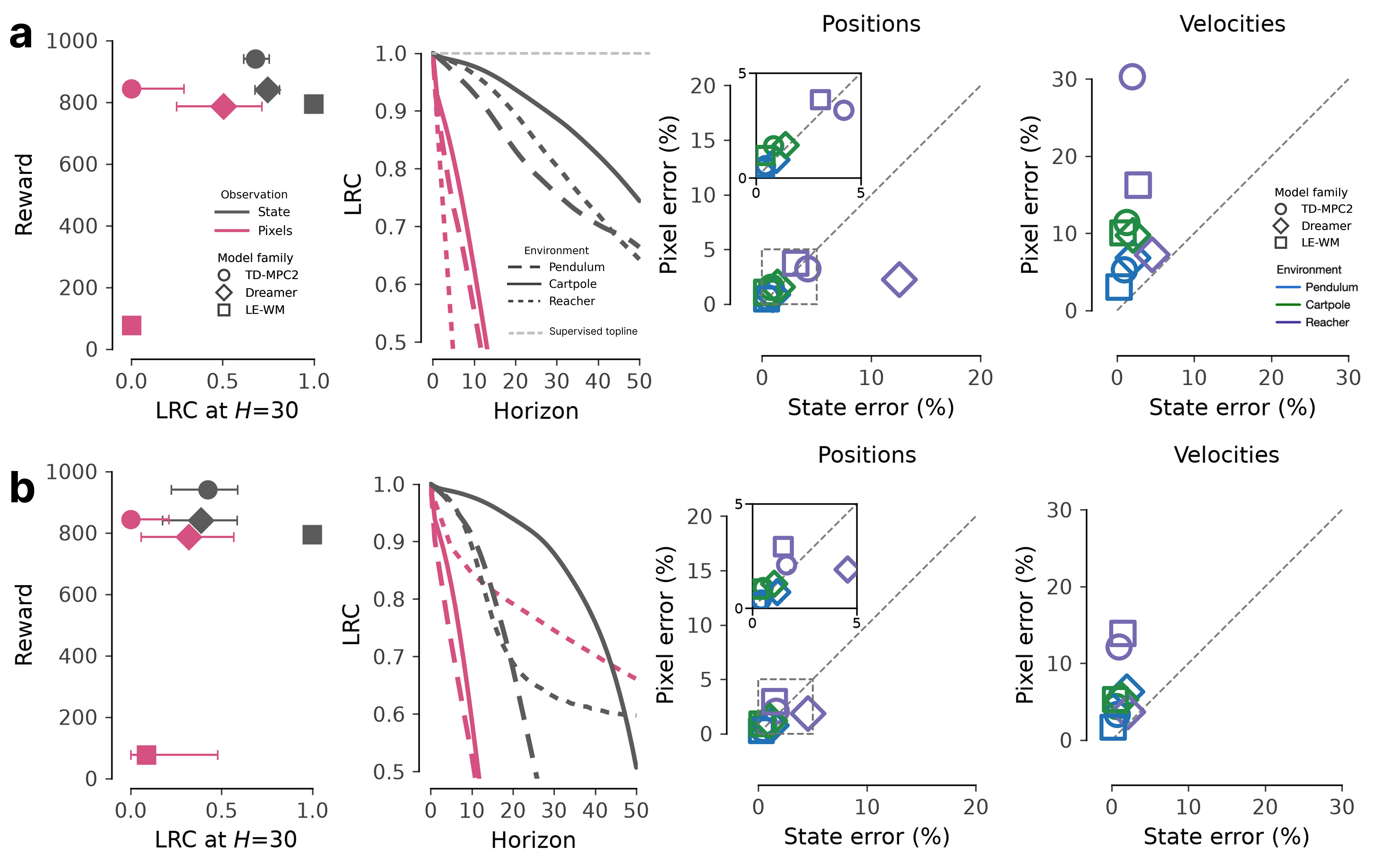}
  \caption{\textbf{Main results on random and own-policy trajectories.}
  Panels as in Figure~\ref{fig:lrc-reward-consistency}, evaluated on random OU data (top) and each model's own trajectories (bottom).}
  \label{fig:own-policy-main-results}
\end{figure}

Figure~\ref{fig:own-policy-main-results} repeats Figure~\ref{fig:lrc-reward-consistency}, with each model evaluated on random OU trajectories (top row) and on its own trajectories (bottom row). The conclusions of the main text do not change. Reward remains dissociated from LRC, and pixel-based TD-MPC2 and DreamerV3 still reach high reward at low LRC at $H=30$. Pixel-based models still decode velocities with larger errors than state-based models, while position errors remain small for both modalities. Averaged across models, LRC over horizons is similar on both datasets, except on Reacher. There, the pixel-based curve no longer collapses within the first few steps on own-policy data. State-based models remain more accurate for roughly the first 10 steps, after which the two curves cross at an LRC of about $0.8$ and pixel-based models stay ahead for the rest of the horizon.

\begin{figure}[H]
  \centering
  \includegraphics[width=\linewidth]{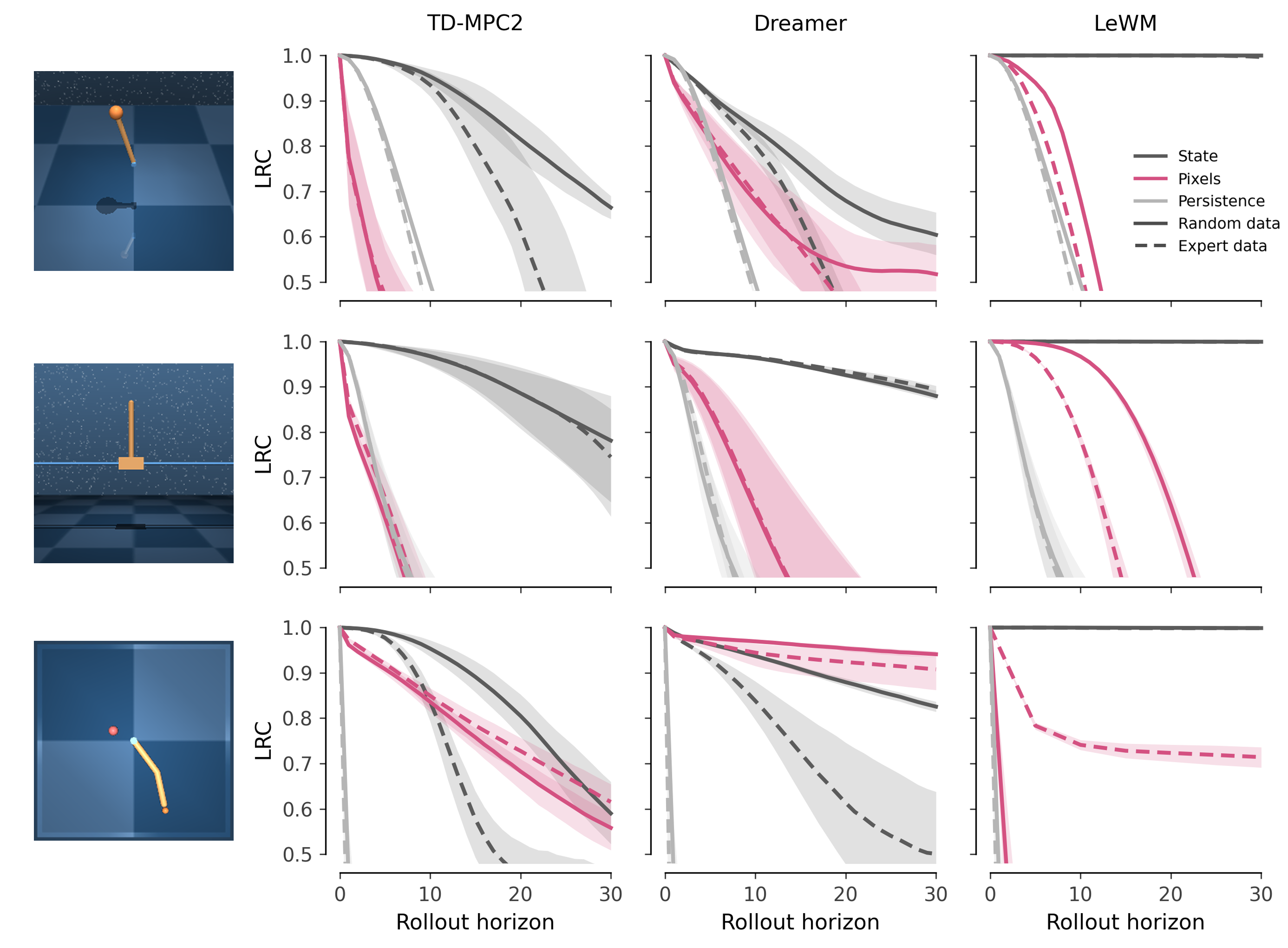}
  \caption{\textbf{LRC over horizons per environment and model family.}
  Rows: Pendulum, CartPole, Reacher. Solid lines: random OU data; dashed lines: own-policy data. Shaded regions show variation across model seeds.}
  \label{fig:lrc-models-environments}
\end{figure}

Figure~\ref{fig:lrc-models-environments} breaks these curves down by model family and environment. In every case except pixel-based LeWM on Reacher, LRC on own-policy data is equal to or lower than on random data. Pixel-based LeWM on Reacher is the exception. Its LRC collapses within a few steps on random data but levels off near $0.7$ on its own trajectories. Because Figure~\ref{fig:own-policy-main-results} averages over model families, this single case accounts for the higher pixel-based Reacher curve there. 

\subsection{Summary}
On trajectories generated by each model's own controller, reward remains dissociated from LRC and pixel-based models remain less accurate than state-based ones. Apart from pixel-based LeWM on Reacher, LRC is equal or lower on own-policy data, so the conclusions of the main text do not depend on evaluating with a random policy.

\section{World-model implementation, training, and evaluation}
\label{app:models}

\subsection{Models and implementation sources}

We evaluate three complementary world-model families: TD-MPC2~\citep{hansentd}, DreamerV3~\citep{hafner2025mastering}, and LeWM~\citep{leworldmodel2026}.
We use the authors' implementations and default configurations as the starting point for all three models, with the modifications described below.
For every model family, observation modality, and environment, we train four independent seeds.
Table~\ref{tab:models} summarizes the resulting experimental matrix.

\begin{table}[H]
  \centering
  \caption{World models evaluated in this work. Latent dimension refers to the size of the activation vector used for all downstream analyses.}
  \label{tab:models}
  \small
  \setlength{\tabcolsep}{4pt}
  \begin{tabularx}{\linewidth}{@{}l l c c >{\raggedright\arraybackslash}X@{}}
    \toprule
    \textbf{Model} & \textbf{Obs.\ type} & \textbf{Seeds} & \textbf{$d_z$} & \textbf{Notes} \\
    \midrule
    \multirow{2}{*}{TD-MPC2}
                    & state  & 4 & 512  & Single-task, per-environment \\
                    & pixels & 4 & 512  & Single-task, per-environment \\
    \addlinespace
    \multirow{2}{*}{DreamerV3}
                    & state  & 4 & 1536 & RSSM posterior: stoch (1024) + deter (512) \\
                    & pixels & 4 & 5120 & Same architecture, visual encoder \\
    \addlinespace
    \multirow{2}{*}{LeWM}
                    & state  & 4 & 192  & Our state-input implementation \\
                    & pixels & 4 & 192  & Original LeWM, $224{\times}224$ frames \\
    \bottomrule
  \end{tabularx}
\end{table}

\paragraph{TD-MPC2.}
We use the official TD-MPC2 implementation (\url{https://github.com/nicklashansen/tdmpc2}) with a 512-dimensional latent and train a separate model for each environment and observation modality.

\paragraph{DreamerV3.}
We use the PyTorch implementation of DreamerV3 (\url{https://github.com/NM512/dreamerv3-torch}) with its standard recurrent state-space model (RSSM).
The state-input model has a 1024-dimensional categorical stochastic state and a 512-dimensional deterministic state, making the total latent representation 1536 dimensional.

\begin{figure}[H]
  \centering
  \includegraphics[width=0.6\linewidth]{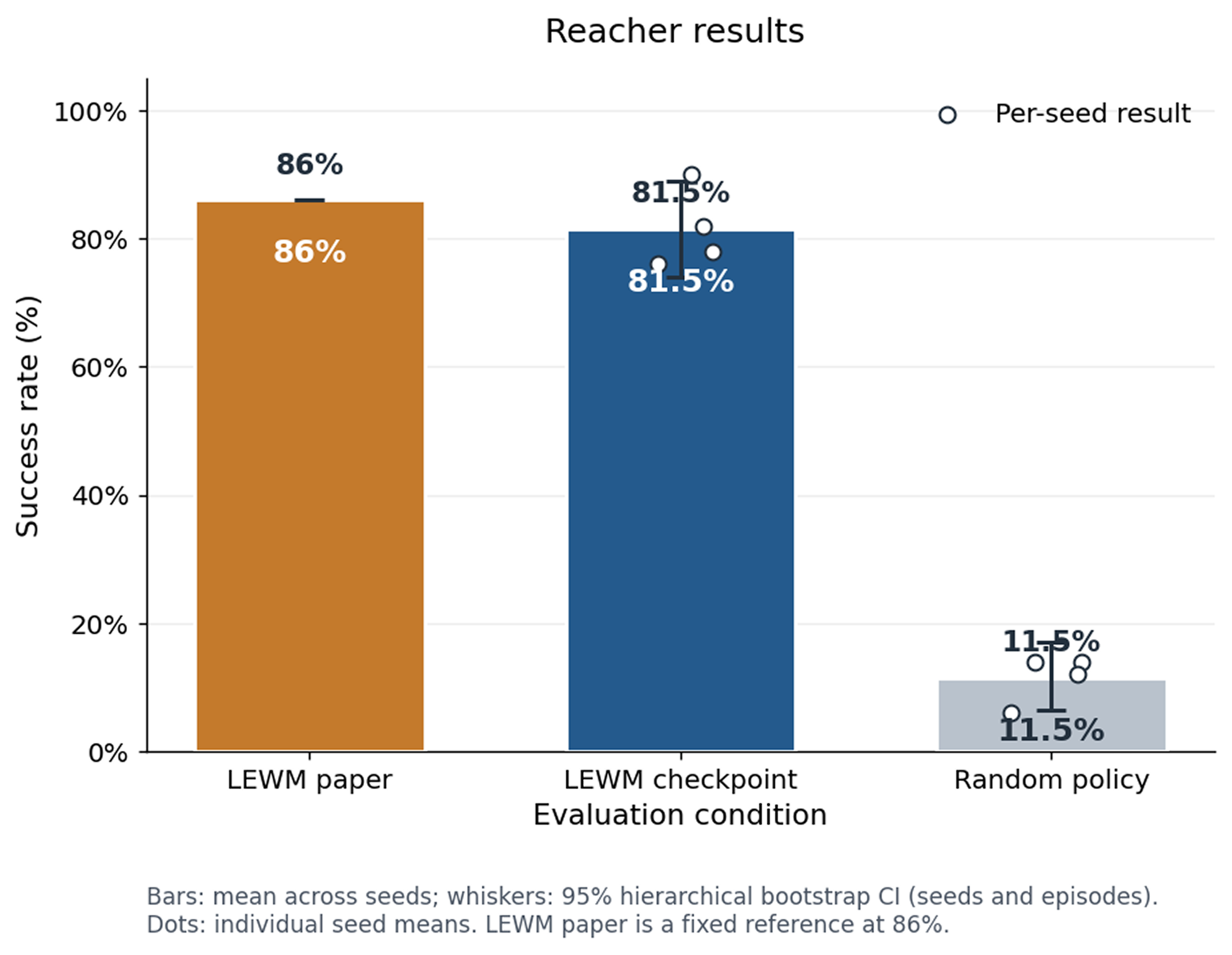}
  \caption{\textbf{Reproduction of LeWM performance on Reacher.}
  The original study reports an 86\% success rate for pixel-based LeWM. Our reproduced checkpoints achieve 81.5\%, compared with 11.5\% for a random policy. Bars show means across four seeds, dots show individual seed means, and whiskers show 95\% hierarchical bootstrap confidence intervals over seeds and episodes. The published result is included as a fixed reference without uncertainty.}
  \label{fig:lewm-reproduction}
\end{figure}

\paragraph{LeWM.}
For pixels, we use the official LeWM implementation (\url{https://github.com/lucas-maes/le-wm}) with $224{\times}224$ inputs and a three-frame context.
Our reproduced Reacher checkpoints achieve an 81.5\% success rate, compared with the 86\% reported by the original study (Figure~\ref{fig:lewm-reproduction}). 
Because the original code does not support state inputs, we implement a state-based variant with a Mish MLP encoder ($d_s \rightarrow 512 \rightarrow 192$) and an action-conditioned residual MLP predictor ($384 \rightarrow 2048 \rightarrow 192$). It operates on a single state vector and uses the same one-step objective as pixel LeWM.
For all downstream analyses, we use the TD-MPC2 encoder output, the LeWM encoder output, and the DreamerV3 RSSM posterior feature vector, concatenating its deterministic state with the softmax probabilities of its categorical stochastic state rather than sampled categories.

\subsection{Training protocol}
\label{app:training}

\paragraph{TD-MPC2 and DreamerV3.}
We train both task-driven models for up to one million environment steps using their original optimization settings and without per-task hyperparameter tuning.
We retain checkpoints after episode reward has converged to a stable high-return regime.

\paragraph{LeWM.}
For each environment, LeWM is trained on 2,000 collected episodes (200,000 transitions): 1,000 from the random OU policy and 1,000 from the TD-MPC2 expert policy.
The same trajectories are used for the state and pixel variants through their aligned state vectors and rendered frames.
We train all LeWM variants for 50 epochs and sweep the SigReg regularization coefficient $\lambda$ \citep{balestriero2025lejepa,leworldmodel2026}, selecting the values that give the best combination of training-loss convergence.
The selected weights are $\lambda=10^{-5}$ for all state-input models, $\lambda=0.001$ for pixel CartPole, and $\lambda=0.1$ for pixel Pendulum and Reacher.
Pendulum and CartPole use a frame skip of 1, whereas Reacher uses a frame skip of 5 following the original LeWM setup. Prediction and SigReg losses converge within the training budget.

\subsection{Task-return evaluation}
\label{app:reward-eval}

\paragraph{TD-MPC2 and DreamerV3.}
For the task-driven models, we evaluate the controller learned jointly with each world model.
TD-MPC2 is run in deterministic evaluation mode, whereas DreamerV3 samples from its learned actor distribution.
For each model seed, we report native environment return over 100 episodes of 500 environment steps.

\begin{figure}[H]
  \centering
  \includegraphics[width=0.6\linewidth]{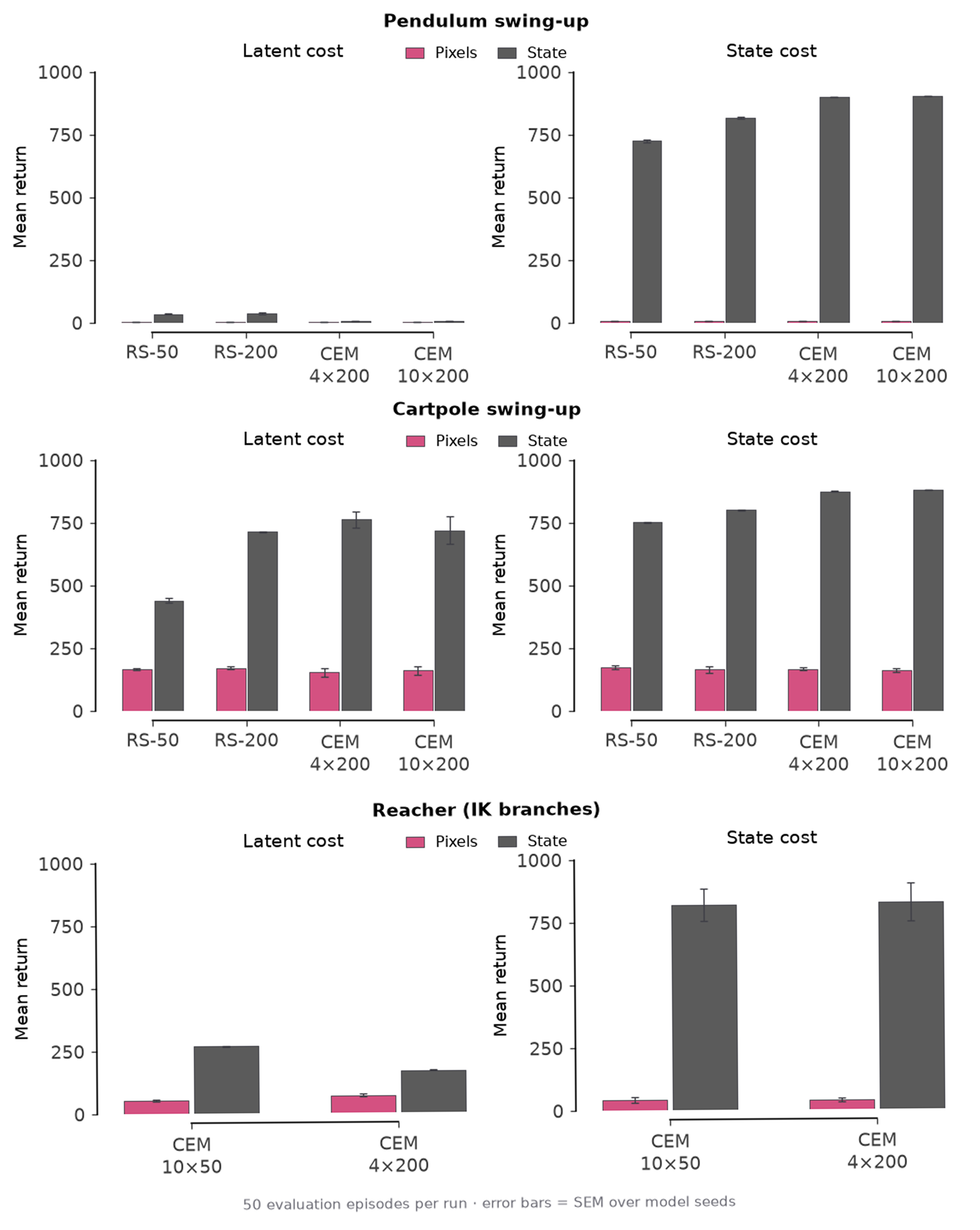}
  \caption{\textbf{LeWM cost and planner comparison.}
  Terminal latent cost and decoded state cost for pixel-input (pink) and state-input (gray) LeWM.
  Planners are random shooting with 50 or 200 candidates and CEM with 4 or 10 iterations of 200 candidates.
  Reacher compares two CEM settings.
  Reported LeWM returns use the state cost and the planner with the highest reward.
  Bars are means over model seeds; error bars are the standard error. Each bar uses 50 evaluation episodes.}
  \label{fig:lewm-controller-sweep}
\end{figure}

\paragraph{LeWM.}
Because LeWM is trained without a policy or reward, we evaluate it with post-hoc model-predictive control.
The original LeWM controller scores an action sequence by the distance between the goal embedding and only the final predicted latent.
Figure~\ref{fig:lewm-controller-sweep} compares that latent cost with a decoded state cost across planners.
The state cost passes each predicted latent through a frozen MLP trained on the aligned simulator states in the LeWM training data, then sums the weighted squared distance to the goal over the horizon.
It attains higher return than the latent cost, so we use it for every environment and both observation modalities.
Among the planners in Figure~\ref{fig:lewm-controller-sweep}, we use the one that gives the highest reward. Returns are the sum of the environment's native rewards over 500-step episodes. For each of the four model seeds we evaluate 50 episodes with each of three planner seeds. 

The state cost for each environment is:

\begin{itemize}
    \item \textbf{Pendulum (swingup):} The goal is the stationary upright state. The cost weights the sine and cosine of the pole angle by 1 and angular velocity by 0.05.
    \item \textbf{CartPole (swingup):} The goal is the stationary upright pole with the cart centered on the rail. The cost assigns unit weight to cart position and to the sine and cosine of the pole angle, and zero weight to both velocities.
    \item \textbf{Reacher (hard):} Two joint configurations can place the fingertip on a reachable target. We solve the analytic inverse kinematics for the elbow-up and elbow-down goals and roll out toward each. The cost weights the two joint angles by 1, the two-dimensional fingertip-to-target displacement by 100, and joint velocities by 0. The return for an episode is the average native Reacher return across the two branches.
\end{itemize}

\subsection{Planning with TD-MPC2's frozen latent dynamics with MPC}
\label{app:tdmpc2-mpc}

We solve each task with model predictive control on TD-MPC2's frozen latent
dynamics. We use pendulum swing-up, cartpole swing-up and reacher (hard), with
models trained either from proprioceptive state or from pixels (the 3 most
recent $64\times64$ RGB frames), and 3 training seeds per configuration. At
every environment step, the planner encodes the current observation and rolls
the latent dynamics forward $H$ steps for each candidate action sequence. To
define the cost, we fit a small MLP probe $g$ that maps latents to the
ground-truth physical state, using random-policy rollouts. Each predicted latent
is decoded with $g$, and the sequence is scored by its weighted squared
distance to a goal state, summed over the horizon:
\begin{equation}
  c(a_{t:t+H-1}) = \sum_{h=1}^{H} \sum_{d} w_d \left( g(z_{t+h})_d - s^{\star}_d \right)^2 .
\end{equation}
The goal $s^{\star}$ is the pole-upright, zero-velocity state for the swing-up
tasks (with the cart centred on cartpole). For reacher, it is the
inverse-kinematics configuration that puts the fingertip on the target. The
weights $w$ cover pole angle and angular velocity for pendulum ($1, 1, 0.05$),
cart position and pole angle for cartpole ($1, 1, 1$), and joint angles plus
fingertip-to-target offset for reacher ($1$ and $100$). TD-MPC2's reward and
value heads are not used.
The cost is minimised with the cross-entropy method (CEM), warm-started from
the previous plan. Swing-up uses $300$ samples $\times$ $30$ iterations with
$30$ elites; reacher uses $50$ samples $\times$ $10$ iterations with $5$
elites. Only the first action is executed before replanning. The one
experimental variable is the planning horizon $H$, in agent steps ($2$
simulator steps each): $H \in \{1, 3, 5, 10, 20, 40, 50, 75\}$ for swing-up and
$H \in \{1, 3, 5, 10, 20, 40\}$ for reacher. We compare against two references.
One is TD-MPC2's native controller: MPPI with $H = 3$, $512$ samples, $6$
iterations, policy-guided samples and a learned terminal value. The other is
uniform random actions. Every condition runs $20$ episodes of $500$ agent steps
(maximum return $\approx 1000$) from identical start states. We report the mean
over the $3$ training seeds of each seed's mean return, with $\pm 1$ standard
deviation across seeds.

\begin{figure}[H]
  \centering
  \includegraphics[width=\linewidth]{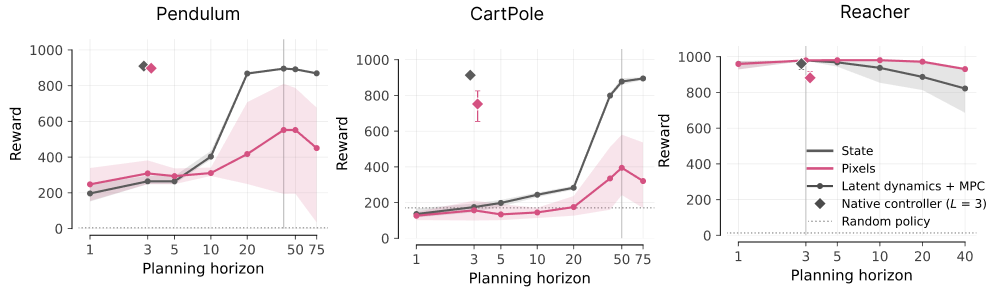}
  \caption{\textbf{Planning horizon and reward for TD-MPC2 with its latent dynamics alone.}
  Reward of MPC on the latent dynamics with a decoded state cost, for state-input (gray) and pixel-input (pink) TD-MPC2, as a function of the planning horizon. Diamonds show the native controller, which plans over $L=3$ steps. The vertical line marks the horizon reported in Figure~\ref{fig:tdmpc2-mpc}, and the dotted line marks a random policy. Shaded regions show variation across model seeds.}
  \label{fig:tdmpc2-mpc-horizon}
\end{figure}

Figure~\ref{fig:tdmpc2-mpc-horizon} shows return across planning horizons $H$. Without the terminal value function, Pendulum and CartPole yield near-random return at short horizons ($H \le 10$) and require $H \ge 20$--$50$ to swing up: at those horizons, state-based TD-MPC2 recovers native-level return ($\approx 900$), whereas pixel-based TD-MPC2 plateaus much lower ($\approx 550$ on Pendulum, $\approx 400$ on CartPole) as rollout errors compound. On Reacher, a greedy one-step lookahead ($H=1$) already suffices ($\approx 950$ return) and pixel LRC remains much higher (Figure~\ref{fig:lrc-models-environments}), so both modalities solve the task across horizons.

\section{Probing details}
\label{app:probing}

\paragraph{Data and targets.}
Probing is run on activation datasets extracted from the random-policy and expert-policy episodes described in Appendix~\ref{app:envs_data}.
Each sample is a full episode containing latent activations $\vz_t$ and aligned ground-truth simulator state $\vs_t$. For each model we train one scalar probe per state coordinate $s_t^{(j)}$, $j \in \{1,\dots,d_s\}$.

\paragraph{Probe families.}
We use three probe configurations:
\begin{itemize}
  \item \textbf{Linear probe}: ordinary least-squares regression from a single latent vector $\vz_t$.
  \item \textbf{Nonlinear probe}: an MLP regressor with hidden sizes $(128,64)$, ReLU activations, $L_2$ penalty $\alpha{=}10^{-4}$, learning rate $10^{-3}$, early stopping (patience 20, validation fraction $0.1$), and a maximum of 300 epochs.

  \item \textbf{Windowed nonlinear probe}: the same MLP architecture applied to a concatenated trailing window of three consecutive latent vectors $[\vz_{t-2};\, \vz_{t-1};\, \vz_t]$.
    This probe is used only for LeWM with pixel inputs, whose latent representation corresponds to a short visual context window rather than a single-step state.
\end{itemize}
Inputs to all probes are standardized feature-wise (zero mean, unit variance).
Targets are kept in physical units and are not standardized.

\paragraph{Shuffled-target control.}
For each environment and physical-state variable, we train a shared control probe that takes the ground-truth state as input and predicts target labels whose correspondence to the inputs has been randomly shuffled. The control uses the same nonlinear architecture and episode-level split as the primary probes. We evaluate it against the original, unshuffled targets in the held-out test episodes. This control is trained independently of any world model, and the same control error is used for every model family, observation modality, and model seed within the corresponding environment and state variable.

\paragraph{Train/test protocol.}
We split at the \emph{episode level}: 80\% of episodes are used for training and 20\% for testing. This prevents leakage across temporally adjacent samples from the same rollout.

\paragraph{Reported metrics.}
We evaluate probes using the mean absolute error (MAE) in physically interpretable units: degrees for angular variables, centimeters for positions, and native units (m/s or deg/s) for velocities. For angular variables, separate probes predict $\cos\theta$ and $\sin\theta$. We reconstruct the angle as $\hat{\theta}=\operatorname{atan2}(\widehat{\sin\theta},\widehat{\cos\theta})$ and compute its error using the shortest circular distance to the target angle. The shuffled-target control undergoes the same reconstruction and evaluation.

For model $m$ and physical-state variable $j$, we define error relative to the shared shuffled-target control as
\begin{equation}
  \epsilon_{m,j}
  =
  100\,
  \frac{\operatorname{MAE}_{m,j}}
       {\operatorname{MAE}^{\mathrm{control}}_j}.
\end{equation}
Thus, $\epsilon_{m,j}=100\%$ denotes the error of the shuffled-target control, values below 100\% indicate that the model's latent representation supports better decoding than the control, and values above 100\% indicate worse performance. All metrics are computed on the held-out 20\% of episodes. We report means and bootstrap confidence intervals across seeds.

\begin{figure}[H]
  \centering
  \includegraphics[width=0.8\linewidth]{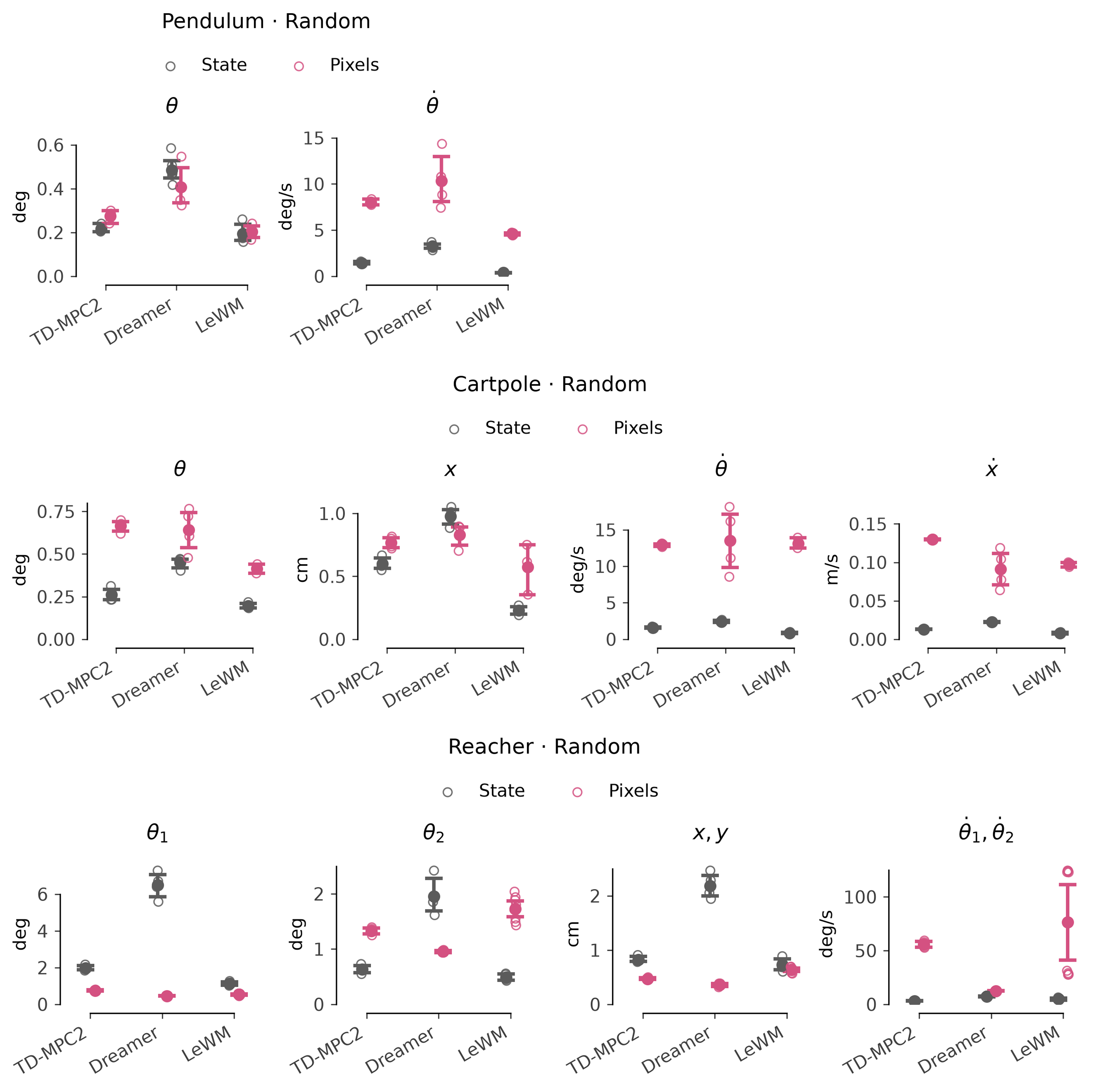}
  \caption{\textbf{State-decoding errors in physical units.}
  Held-out nonlinear-probe errors for models trained from ground-truth states or pixels, reported separately for Pendulum, CartPole, and Reacher. Panels separate the positional and velocity coordinates of each environment and report errors in degrees, centimeters, meters per second, or degrees per second as appropriate. Colored markers show individual model seeds; black markers and error bars summarize performance across seeds.}
  \label{fig:state-decoding-physical}
\end{figure}

\begin{figure}[H]
  \centering
  \includegraphics[width=\linewidth]{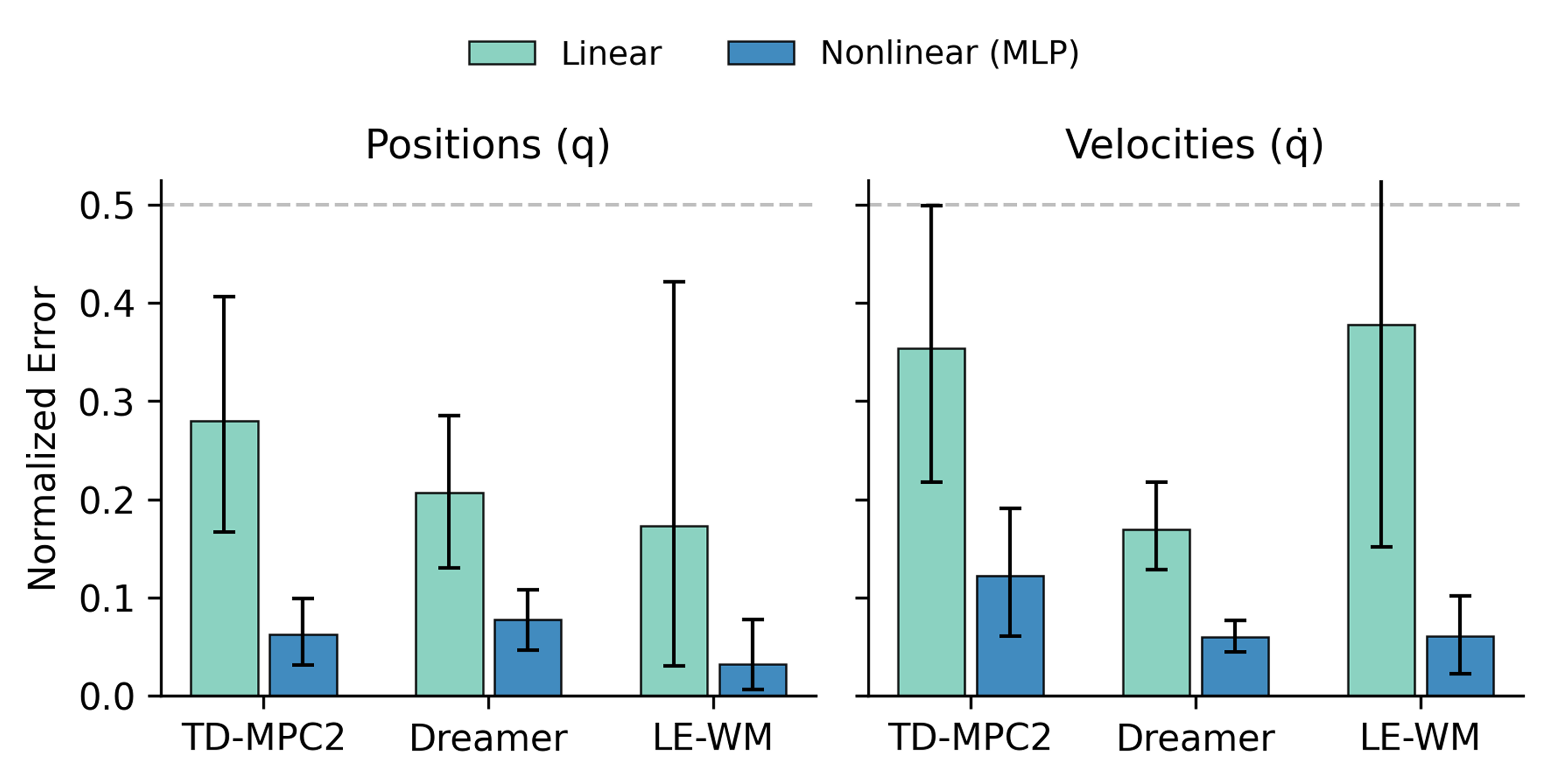}
  \caption{\textbf{Linear and nonlinear state decoding.}
  Normalized decoding error for positional variables (left) and velocities (right), averaged across environments for each model family. Bars compare linear probes with nonlinear MLP probes; error bars show variation across environments. The dashed line marks an error equal to 50\% of the shuffled-target control.}
  \label{fig:state-decoding-probes}
\end{figure}

\section{Controlled-intervention details}
\label{app:interventions}

\subsection{Nonlinear state mixing}
\label{app:state-mixing-exp}
We perform this intervention on CartPole, applying a fixed mixing network to its five-dimensional state before the encoder of each world model.
Depth zero is the identity; depths one to three compose bias-free $5{\times}5$ linear maps with LeakyReLU activations (negative slope $0.2$) between consecutive maps.
Following the construction of \citet{hyvarinen2019nonlinear}, each weight matrix is sampled with column-normalized entries and retained only if its condition number lies below the 25th percentile of 10,000 random matrices.
The mixer uses seed 42, remains frozen, and is shared across model seeds at a given depth.
We train a separate TD-MPC2, DreamerV3, and LeWM model from scratch for every depth using the corresponding primary training protocol; the standard pixel models are included only as a reference.

\subsection{Velocity observability}
\label{app:velocity-exp}
We perform the velocity intervention on pixel-based CartPole models using the ground-truth cart and pole velocities $(\dot{x},\dot{\theta})$ without normalization.
For each model, a small MLP maps the two velocities to a 16-dimensional feature, which is concatenated with the visual embedding and projected back to the model's standard representation size.
The resulting velocity-fused encoder, latent dynamics, and controller (for TD-MPC2 and DreamerV3) are trained jointly from scratch; velocity is supplied only while encoding observations and is not provided during subsequent open-loop latent prediction.
TD-MPC2 and LeWM are trained under all four combinations of one or three input frames and with or without velocity.
The DreamerV3 encoder processes one $64{\times}64$ image at each timestep, while temporal context is maintained by the recurrent RSSM.
In the $+v$ condition, the ground truth two-dimensional \texttt{velocity} observation is mapped through a two-layer MLP to a 16-dimensional feature, concatenated with the CNN image embedding, and projected back to the original embedding size before entering the RSSM.
We use four seeds for TD-MPC2 and DreamerV3 and three seeds for LeWM, and otherwise retain the primary training and evaluation protocols.

Figure~\ref{fig:lewm-controller-ablations} shows the same comparison for the CartPole velocity and mixing checkpoints. Both use the state cost and the highest-reward planner from Appendix~\ref{app:reward-eval}.

\begin{figure}[H]
  \centering
  \includegraphics[width=\linewidth]{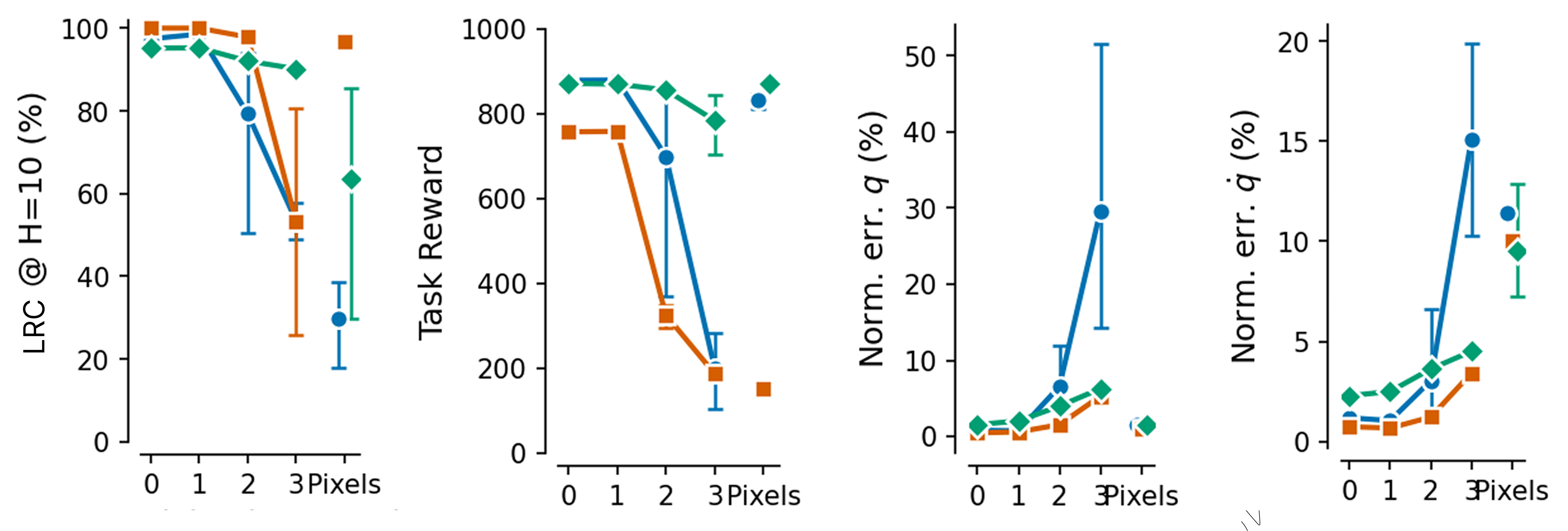}
  \caption{\textbf{Nonlinear state mixing across world-model families.}
  LRC at $H=10$, task reward, normalized position error, and normalized velocity error on CartPole as the mixing depth increases from zero to three layers. Results are shown for TD-MPC2 (blue circles), LeWM (orange squares), and DreamerV3 (green diamonds); \emph{Pixels} denotes each model's standard pixel-input condition. Error bars show variation across model seeds.}
  \label{fig:nonlinear-mixing-all-models}
\end{figure}

\begin{figure}[H]
  \centering
  \includegraphics[width=\linewidth]{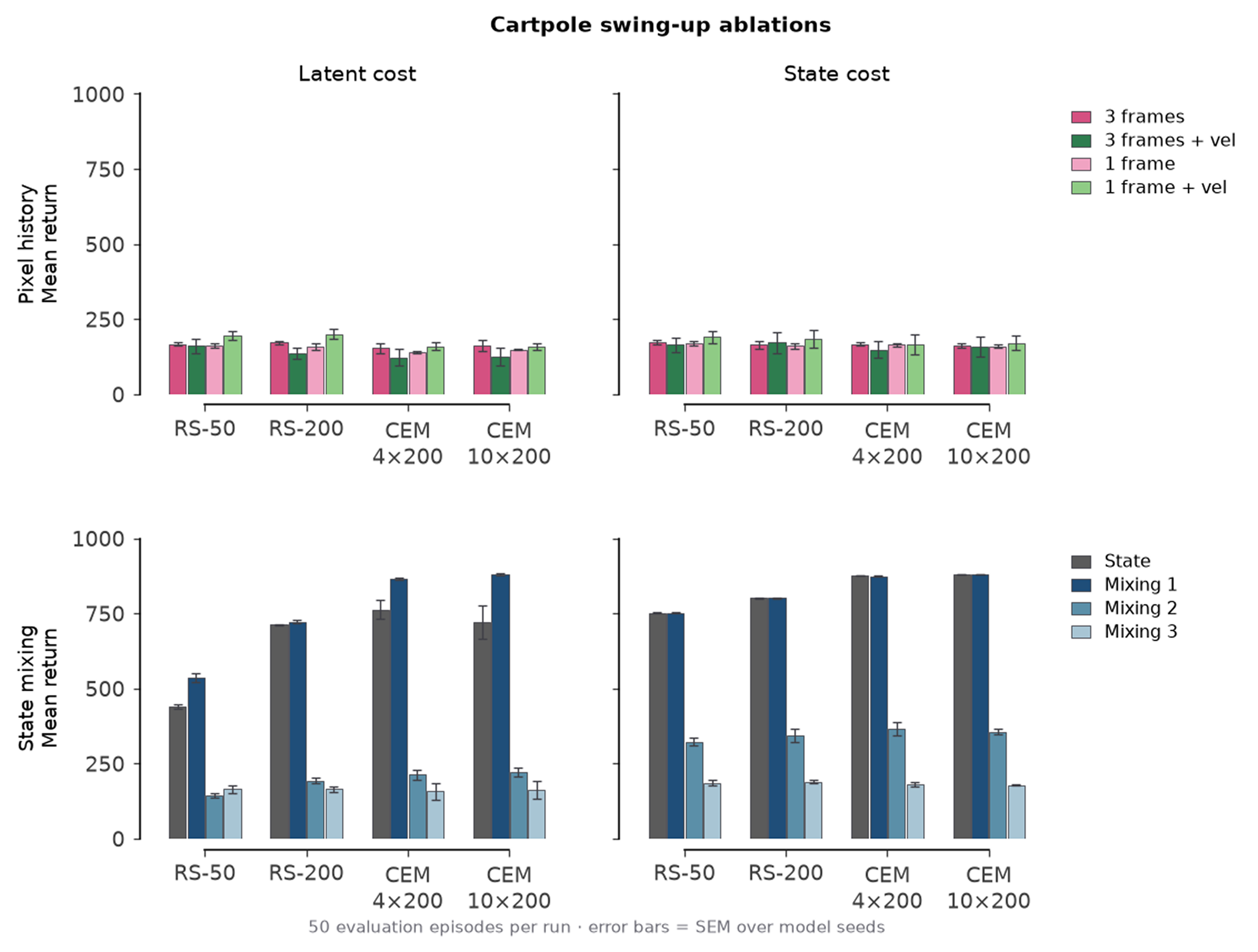}
  \caption{\textbf{LeWM controllers on the CartPole interventions.}
  Top: pixel models with one or three frames, with and without ground-truth velocity.
  Bottom: state-input models after zero to three nonlinear mixing layers.
  Columns compare the terminal latent cost with the decoded state cost across the planners in Figure~\ref{fig:lewm-controller-sweep}.
  Bars are means over model seeds; error bars are the standard error. Each bar uses 50 evaluation episodes.}
  \label{fig:lewm-controller-ablations}
\end{figure}

\section{Synthetic perturbations of the faithful condition}
\label{app:faithful-perturbations}

Figure~\ref{fig:hypotheses} uses an oracle CartPole planner. We corrupt either the dynamics or the encoder and measure how reward, LRC, and state decoding move. There is no learned world model.

\paragraph{Task and controller.}
We use DeepMind Control Suite CartPole swing-up~\citep{tunyasuvunakool2020}.
The planner is random-shooting MPC: at each decision it samples $N=50$ action sequences of length $H=10$ over $\{-1,0,1\}$, simulates each sequence for $k=5$ transition steps per horizon step, selects the lowest-cost sequence, and executes its first action in the true environment.
Episodes last 500 environment steps.
The planning cost is weighted mean squared error to the upright goal in the canonical state $[x,\cos\theta,\sin\theta,\dot x,\dot\theta]$, with unit weight on the position and angle coordinates and zero weight on both velocities.
The internal model is the ground-truth CartPole transition $T$, not a learned $\hat{\rvf}$.
Reported reward is the sum of the environment's native rewards over the episode.

\paragraph{Conditions.}
The three columns of Figure~\ref{fig:hypotheses} are:
\begin{itemize}
  \item \textbf{A.} Observations are the canonical state $\vs$. The encoder is the identity. The planner uses the noisy oracle $T+\varepsilon$.
  \item \textbf{B.} Observations are a fixed invertible mix $\vy=\rvg(\vs)$. The planner acts in $\vy$ with conjugated dynamics $\vy'=\rvg\bigl(T(\rvg^{-1}(\vy),\va)+\varepsilon\bigr)$.
  \item \textbf{C.} Observations are the same $\vy$. The encoder is a noisy inverse $\rvg^{-1}+\varepsilon$. The planner uses the clean oracle $T$.
\end{itemize}

\paragraph{Encoding.}
Column A uses the identity on the five-dimensional canonical state.
Columns B and C use the depth-3 mixer of Appendix~\ref{app:state-mixing-exp}, frozen at seed 42.
The mixer is not varied in this sweep.

\paragraph{Noise.}
Both dynamics and encoder corruptions use relative Gaussian noise: we add $\varepsilon\,\sigma\odot\xi$ in canonical state units, where $\xi\sim\mathcal{N}(0,I)$ and $\sigma$ is a fixed per-dimension scale.
After dynamics noise we renormalize $(\cos\theta,\sin\theta)$ to the unit circle.
The scale is $\varepsilon\in\{0,0.01,\ldots,0.10,0.20,0.30,0.40,0.50\}$ (15 values).
Columns A and B inject noise in the oracle transition and leave the inverse map clean.
Column C injects noise on $\rvg^{-1}$, perturbing all state components, and sets dynamics noise to zero.
Only one noise channel is active in each condition.

\paragraph{Replication.}
We run three independent MPC rollouts with seeds $\{0,1,2\}$ per condition and $\varepsilon$, and average the metrics.
The mixer seed remains 42.
Decoder diagnostics use deterministic sample streams derived from each MPC seed for the train/eval split.

\paragraph{Metrics.}
The $y$-axis of Figure~\ref{fig:hypotheses} is the cumulative environment reward of the 500-step rollout; point color is $\varepsilon$.
The three $x$-axes are:
\begin{itemize}
  \item \textbf{Linear $R^2$.} Ridge regression from the planning representation to the canonical state on 10{,}000 held-out states. Column A uses the identity, so this value is set to 1.
  \item \textbf{Nonlinear $R^2$.} The same targets, decoded with an MLP with hidden sizes $(64,64)$.
  \item \textbf{LRC.} The same metric as in the main text (Eq.~\ref{eq:lrc}), evaluated at horizon $H=30$.
\end{itemize}

\section{Analysis on Multi-task TD-MPC2 on 30 DMControl tasks}
\label{app:multitask}

The main analyses use three deliberately simple environments. To test whether the findings extend to a broader range of tasks and embodiments, we apply the main analysis of the paper to a single large multi-task world model.

\paragraph{Model and tasks.}
We analyse the released TD-MPC2 multi-task checkpoint (317M parameters)~\citep{hansentd}, trained jointly on 30 DMControl tasks spanning 11 embodiments. Its encoder and latent dynamics are conditioned on a learned 96-dimensional task embedding, and the latent state has 1376 dimensions. All tasks are analysed with this single model. This model receives proprioceptive state inputs.

\paragraph{Data.}
For each task, we sample 1000 episodes uniformly from that task's episodes in the released TD-MPC2 mt30 dataset. This dataset consists of the replay buffers of the single-task TD-MPC2 agents used to train the multi-task model. We keep the first 100 steps of each episode. These replay buffers mix exploratory and near-expert behaviour.

\begin{figure}[H]
  \centering
  \includegraphics[width=\linewidth]{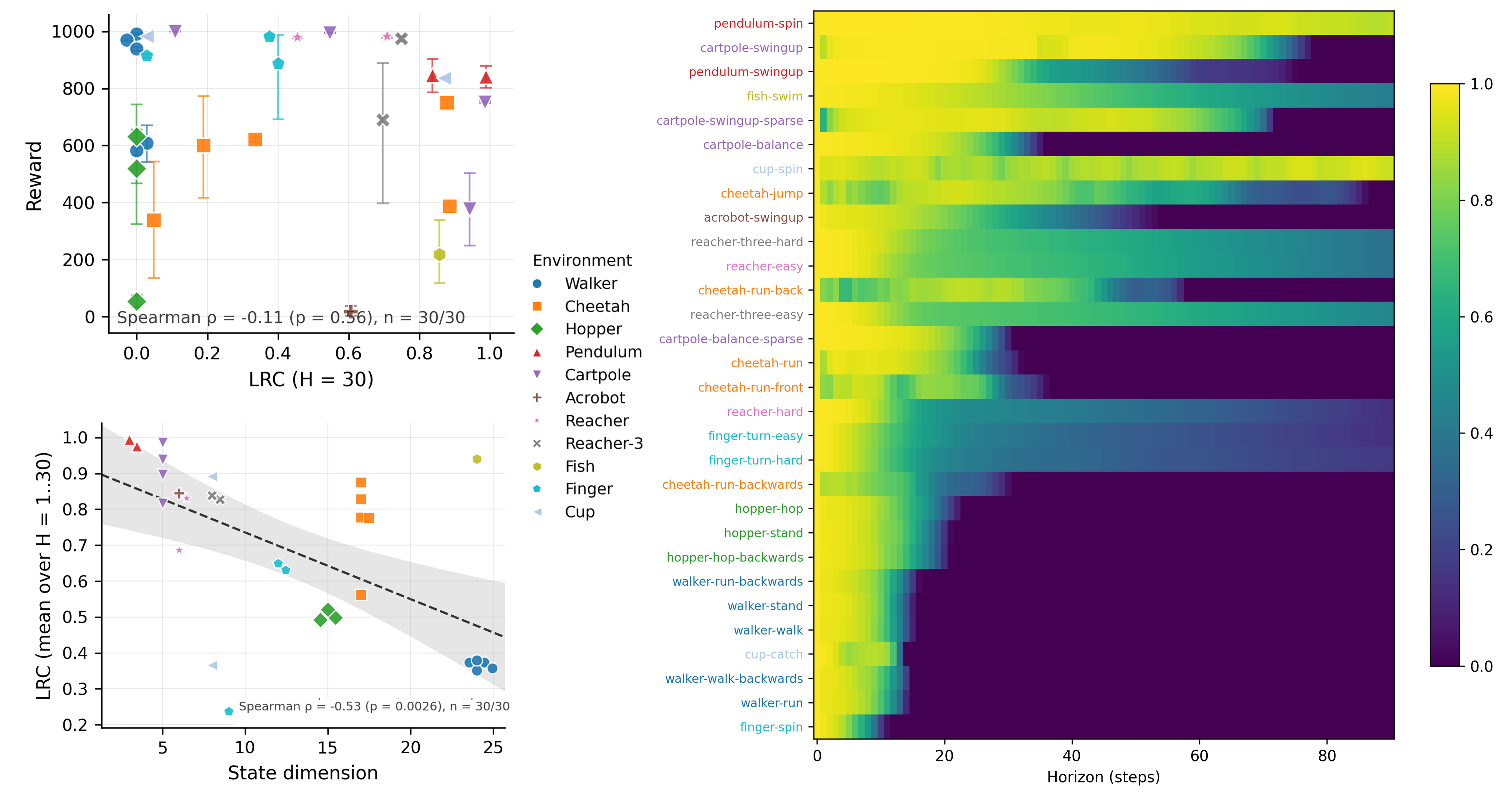}
  \caption{\textbf{Multi-task TD-MPC2 on 30 DMControl tasks.}
  Top left: reward versus LRC at $H=30$ for each task, colored by embodiment, with the Spearman correlation across tasks.
  Bottom left: LRC averaged over horizons $H=1,\ldots,30$ versus the dimension of each task's state, with a linear fit and its confidence band.
  Right: LRC over the rollout horizon for each task and sorted by mean LRC over 30 steps.}
  \label{fig:multitask}
\end{figure}

\paragraph{Results.}
Across the 30 tasks, reward and LRC at $H=30$ are uncorrelated (Spearman $\rho=-0.11$, $p=0.56$; Figure~\ref{fig:multitask}, top left). Several tasks, including most walker tasks, reach near-maximal reward with LRC close to zero. This extends the main finding of the paper beyond our three environments: high reward does not imply accurate latent dynamics. In addition, LRC decreases with the dimension of the task's state (Spearman $\rho=-0.53$, $p=0.0026$; Figure~\ref{fig:multitask}, bottom left). Low-dimensional tasks such as pendulum and cartpole keep accurate rollouts over long horizons, whereas higher-dimensional locomotion tasks such as walker and hopper lose rollout accuracy within a few steps (Figure~\ref{fig:multitask}, right), even though the model receives states rather than pixels.

\end{document}